\documentclass[10pt,twocolumn,letterpaper]{article}

\usepackage[pagenumbers]{cvpr} %

\usepackage{graphicx}
\usepackage{amsmath}
\usepackage{amssymb}
\usepackage{booktabs}
\usepackage{enumitem}
\usepackage{multirow}
\usepackage{amsthm}
\usepackage{amsmath}
\usepackage{mathrsfs}
\usepackage{bbold}
\usepackage{algorithmic}
\usepackage{xcolor}
\usepackage{soul}
\usepackage[ruled, linesnumbered]{algorithm2e}
\usepackage{pifont}
\usepackage{flushend}
\usepackage[accsupp]{axessibility}
\definecolor{britishracinggreen}{rgb}{0.0, 0.26, 0.15}
\definecolor{aoenglish}{rgb}{0.0, 0.5, 0.0}
	
\newcommand{\xmark}{\ding{55}}
\newcommand{\cmark}{\ding{51}}	

\newcommand{\increase}[1]{\scriptsize \textcolor{aoenglish}{$\downarrow$#1} \normalsize}

\usepackage[pagebackref,breaklinks,colorlinks]{hyperref}

\usepackage[capitalize]{cleveref}

\newcommand{\supp}[1]{{\color{blue}  #1}}

\crefname{section}{Sec.}{Secs.}
\Crefname{section}{Section}{Sections}
\Crefname{table}{Table}{Tables}
\crefname{table}{Tab.}{Tabs.}

\def\cvprPaperID{9379} %
\def\confName{CVPR}
\def\confYear{2022}

\begin{document}

\title{\textbf{SPAct}: Self-supervised Privacy Preservation for Action Recognition}

\author{Ishan Rajendrakumar Dave, Chen Chen, Mubarak Shah\\
Center for Research in Computer Vision, University of Central Florida, Orlando, USA\\
{\tt\small ishandave@knights.ucf.edu, \{chen.chen, shah\}@crcv.ucf.edu}
}

\maketitle

\begin{abstract}
    Visual private information leakage is an emerging key issue for the fast growing applications of video understanding like activity recognition. Existing approaches for mitigating privacy leakage in action recognition require privacy labels along with the action labels from the video dataset. However, annotating frames of video dataset for privacy labels is not feasible. Recent developments of self-supervised learning (SSL) have unleashed the untapped potential of the unlabeled data. For the first time, we present a novel training framework which removes privacy information from input video in a self-supervised manner without requiring privacy labels. Our training framework consists of three main components: anonymization function, self-supervised privacy removal branch, and action recognition branch. We train our framework using a minimax optimization strategy to minimize the action recognition cost function and maximize the privacy cost function through a contrastive self-supervised loss. Employing existing protocols of known-action and privacy attributes, our framework achieves a competitive action-privacy trade-off to the existing state-of-the-art supervised methods. In addition, we introduce a new protocol to evaluate the generalization of learned the anonymization function to novel-action and privacy attributes and show that our self-supervised framework outperforms existing supervised methods. Code available at: \url{https://github.com/DAVEISHAN/SPAct}

\vspace{-3mm}

\end{abstract}

\section{Introduction}
\label{sec:intro}
\begin{figure*}[h]
\begin{center}
  \includegraphics[width=0.835\textwidth]{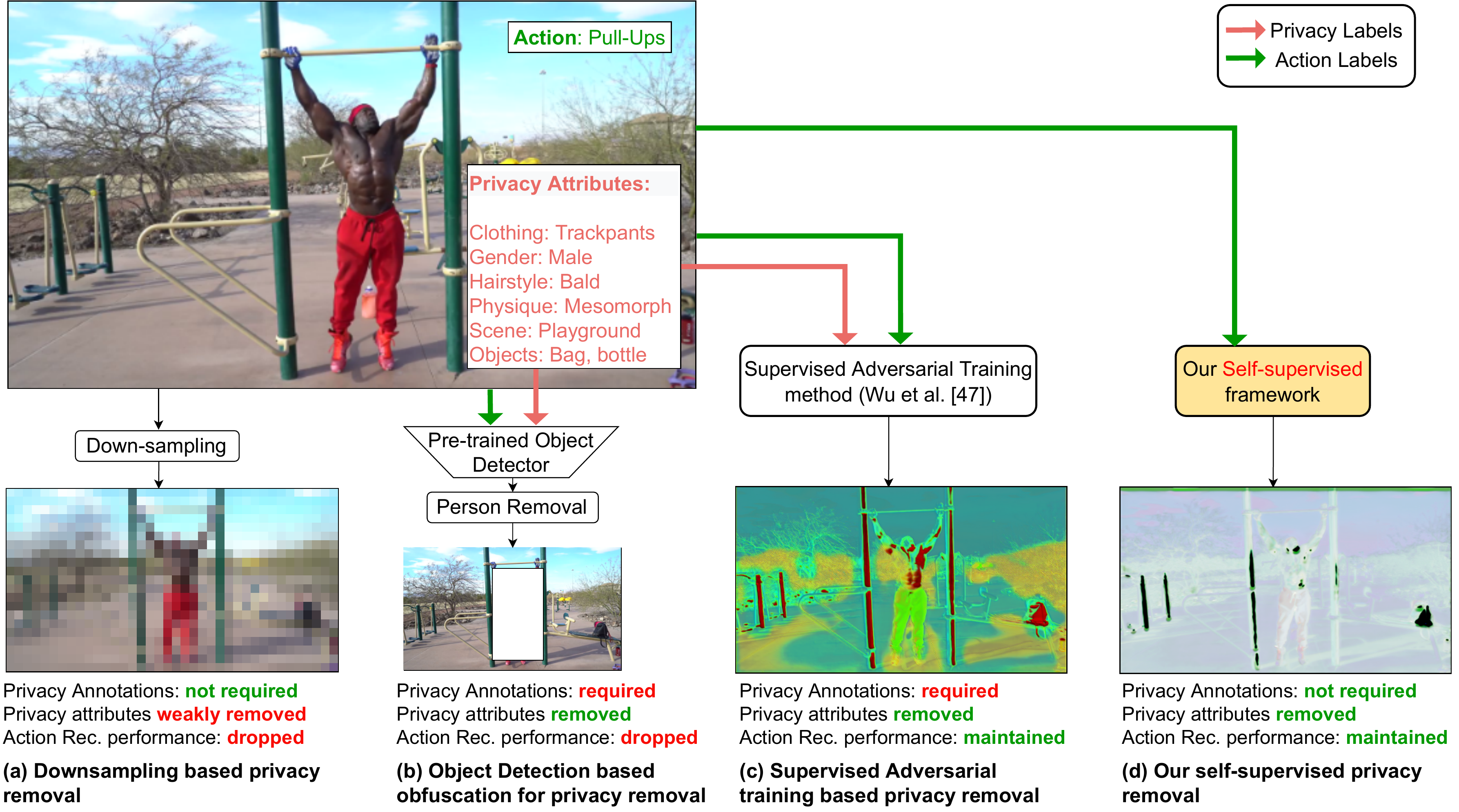}
\end{center}
\vspace{-0.5cm}
\caption{Overview of the existing privacy preserving action recognition approaches. The main goals of a framework include removing privacy information and maintaining action recognition performance at low cost of annotations.}
\vspace{-3mm}

\label{fig:teaser}
\end{figure*}

Recent advances in action recognition have enabled a wide range of real-world applications, \eg video surveillance camera~\cite{cmu2020, rizve2021gabriella, gabv2}, smart shopping systems like \textit{Amazon Go}, elderly person monitor systems~\cite{buzzelli2020vision, zhang2012privacy, liu2020privacy}. Most of these video understanding applications involve extensive computation, for which a user needs to share the video data to the cloud computation server. While sharing the videos to the cloud server for the utility action recognition task, the user also ends up sharing the private visual information like gender, skin color, clothing, background objects etc. in the videos as shown in Fig.~\ref{fig:teaser}. Therefore, there is a pressing need for solutions to privacy preserving action recognition.

A simple-yet-effective solution for privacy preservation in action recognition is to utilize very low resolution videos (Fig.~\ref{fig:teaser}a) ~\cite{ryoo, ishwar, liu2020indoor}. Although this downsampling method does not require any specialized training to remove privacy features, it does not provide a good trade-off between action recognition performance and privacy preservation. 

Another set of methods use pretrained object-detectors to detect the privacy regions and then remove or modify the detected regions using synthesis~\cite{ren2018learning} or blurring~\cite{zhang2021multi} as shown in Fig.~\ref{fig:teaser}b. The detection-based approaches require the bounding-box level annotations for the privacy attributes, and removing the privacy features without an end-to-end learning framework may result in the performance drop of the action recognition task. 

Wu \etal~\cite{wu_tpami} propose a novel approach to remove the privacy features via learning an \textit{anonymization function} through an adversarial training framework, which requires both \textit{action and privacy labels} from the video. Although the method is able to get a good trade-off of action recognition and privacy preservation, it has two main problems. First, it is not feasible to annotate a video dataset for privacy attributes. For instance, Wu \etal~\cite{wu_tpami} acknowledge the issue of privacy annotation time, where it takes immense efforts for them to annotate privacy attributes for even a small-scale (515 videos) video dataset \textit{PA-HMDB}. Second, the learned anonymization function from the known privacy attributes may not generalize in anonymizing the \textit{novel privacy attributes}. For example, in Fig.~\ref{fig:teaser} the learned anonymization function for human-related privacy attributes (\eg gender, skin color, clothing) may still leave other privacy information like scene or background objects un-anonymized.

The performance of the action recognition task depends on the spatio-temporal cues of the input video. Wu \etal~\cite{wu_tpami} show that anonymizing the privacy features like face, gender, etc. in the input video does not lead to any reduction in the action recognition performance. Instead of just focusing on the cues based on the privacy annotations, our goal is twofold: 1) learning an anonymization function that can remove all spatial cues in all frames without significantly degrading action recognition performance; and 2) learning the anonymization function without any privacy annotations. 

Recently, self-supervised learning (SSL) methods have been successfully used to learn the representative features which are suitable for numerous downstream tasks including classification, segmentation, detection, etc. 
Towards our goal, we propose a novel frame-level SSL method to remove the semantic information from the input video, while maintaining the information that is useful for the action recognition task. We show that our proposed \textit{\textbf{S}elf-supervised \textbf{P}rivacy-preserving \textbf{Act}ion recognition (SPAct)} framework is able to anonymize the video without requiring any privacy annotations in the training. 

The learned anonymization function should provide a model-agnostic privacy preservation, hence, we first adopt the protocol from~\cite{wu_tpami} to show the transferability of the anonymization function across different models. \textit{However, there are two aspects in terms of evaluating the generalization ability of the anonymization function, which are overlooked in previous works.}

First, in the real-world scenario, the anonymization function is expected to have \textit{generalization capability with domain shift in action and privacy classes}. To evaluate the generalization capabilities of the anonymization function across novel action and privacy attributes, we propose new protocols. In our experiments, we show that since our model is not dependent on the predefined privacy features like existing supervised methods, and it achieves state-of-the-art generalization across novel privacy attributes.  

Second, prior privacy-preserving action recognition works have solely focused on privacy attributes of humans. In practical scenarios, privacy leakage can happen in terms of scene and background objects as well, which could reveal personal identifiable information. Therefore, \textit{the generalization ability of anonymization function to preserve privacy attributes beyond humans (\eg scene and object privacy)} is of paramount importance as well.
To evaluate such ability, we propose \textit{P-HVU} dataset, a subset of LSHVU dataset~\cite{hvu}, which has multi-label annotations for actions, objects and scenes. Compared to existing same-dataset privacy-action evaluation protocol on PA-HMDB~\cite{wu_tpami}, which consists of only 515 test videos, the proposed P-HVU dataset has about 16,000 test videos for robust evaluation of privacy-preserving action recognition.

\noindent{The contributions of this work are summarized as follows:}
\setlist{nolistsep}
\begin{itemize}
\item We introduce a novel {\em self-supervised} learning framework for privacy preserving action recognition without requiring any privacy attribute labels.
\item On the existing UCF101-VISPR and PA-HMDB evaluation protocols, our framework achieves a competitive performance compared to the state-of-the-art {\em supervised} methods which require privacy labels.
\item %
We propose new evaluation protocols for the learned anonymization function to evaluate its generalization capability across novel action and novel privacy attributes. For these protocols, we also show that our method outperforms state-of-the art supervised methods. Finally, we propose a new dataset split \textit{P-HVU} to resolve the issue of smaller evaluation set and extend the privacy evaluation to non-human attributes like action scene and objects.
\end{itemize}

\section{Related Work}

Recent approaches for the privacy preservation can be categorized in three major groups: (1) Downsampling based approaches; (2) Obfuscation based approaches; and (3) Adversarial training based approaches. An overview of the existing privacy preserving approaches can be seen in Fig.~\ref{fig:teaser}.

Downsampling based approaches utilized a very low resolution input to anonymize the personal identifiable information%
. Chou \etal~\cite{chou2018privacy} utilize low resolution depth images to preserve privacy in the hospital environment. Srivastava \etal~\cite{srivastav2019human} utilize low resolution images to mitigate privacy leakage in human pose estimation. Butler \etal ~\cite{butler2015privacy} use operations like blurring and superpixel clustering to anonymize videos. There are some works ~\cite{ryoo, ishwar, liu2020indoor} utilizing a downsampling based solution for privacy preserving action recognition. An example of anonymization by downsampling is shown in Fig. \ref{fig:teaser}a. Although it is a simple method and does not require privacy-labels for training, one major drawback of the method is its suboptimal trade-off between action recognition and privacy preservation. 

Obfuscation based approaches mainly involve using an off-the-shelf object detector to first detect the privacy attributes and then remove or modify the detected regions to make it less informative in terms of privacy features. An interesting solution is proposed by Ren \etal~\cite{ren2018learning} for anonymizing faces in the action detection utility. They synthesise a fake image in place of the detected face. A similar approach was taken for the video domain privacy by Zhang \etal~\cite{zhang2021multi}, where first the semantic segmentation is employed to detect the regions of interest, which is followed by a blurring operation to reduce the privacy content of a video. Although the obfuscation based methods work well in preserving the privacy, there are two main problems associated with them: (1) there is domain knowledge required to know the regions of interests, and (2) the performance of the utility task is significantly reduced since this approach is not end-to-end and involves two separate steps: private-object detection/segmentation and object removal. %

Recently, Hinojosa \etal~\cite{hinojosa2021learning} tackle the privacy preserving human pose estimation problem by optimizing an optical encoder (hardware-level protection) with a software decoder. In addition, some more work focus on \textit{hardware level} protection in the image based vision systems~\cite{wang2019privacy,pittaluga2015privacy, pittaluga2016pre, jia2013using}, however, they are not within scope of this paper. 

Pittaluga \etal~\cite{pittaluga2019learning} and Xiao \etal~\cite{xiao2020adversarial} propose adversarial optimization strategies for the privacy preservation in the images. Authors in ~\cite{wu_eccv,wu_tpami} introduce a novel adversarial training framework for privacy preserving action recognition. Their framework adopts a minimax optimization strategy, where action classification cost function is minimized, while privacy classification cost is maximized. Their adversarial framework remarkably outperforms prior works which are based on obfuscation and downsampling. 

Recently, self-supervised learning (SSL) based methods have demonstrated learning powerful representations for images~\cite{simclr, moco, byol, zbontar2021barlow} and
videos~\cite{Feichtenhofer_2021_CVPR, videomoco, motionfit, jenni2021time, tclr, iccv21qian, ranasinghe2021self}, which are useful for multiple image and video understanding downstream tasks. In this paper, we propose self-supervised privacy preservation method. Instead of using a privacy classifier to remove only the privacy attributes from the input data like~\cite{wu_tpami}, our approach is to remove \textit{all spatial semantic} information from the video, along with keeping the useful \textit{utility action recognition information} by training an anonymization function in an minimax optimization manner. 
To the bes of our best knowledge, there is no other \textit{self-supervised} privacy preserving 
action recognition method, which learns in an \textit{end-to-end fashion}, without requiring \textit{privacy labels}. %

\begin{figure*}
\begin{center}
  \includegraphics[width=0.87\textwidth]{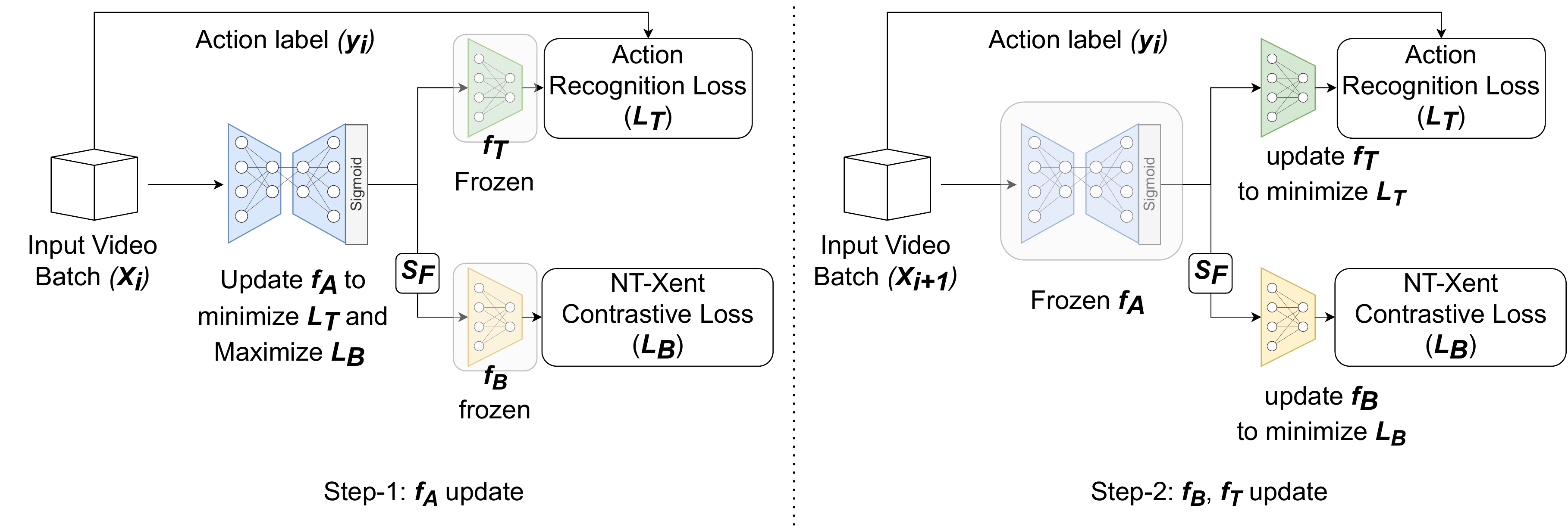}
\end{center}
\vspace{-0.5cm}
\caption{\textbf{Minimax optimization in the proposed SPAct framework}. $f_{A}$ is anonymization function, $f_{T}$ is 3D-CNN based action classifier, $f_{B}$ is 2D-CNN based self-supervised learning model, and $S_{F}$ is temporal sampler. Details of each component can be found in Sec~\ref{sec:framework}. We first initialize $f_{A}$ to identity function and $f_{T}$ and $f_{B}$ to pretrained checkpoints optimized on raw video. The proposed minimax optimization strategy is an iterative process including two-steps per iteration. In the left figure, we first update the weights of $f_{A}$ to minimize action classification loss, $L_{T}$, and maximize NT-Xent contrastive self-supervised loss~\cite{simclr} $L_{B}$, keeping $f_{T}$ and $f_{B}$ frozen. After that as shown in the right figure, for the next batch of videos, we keep $f_{A}$ frozen and update parameters of $f_{T}$ and $f_{B}$ to minimize $L_{T}$ and $L_{B}$, respectively. For more details see Sec~\ref{sec:minimax}.}
\vspace{-4mm}
\label{fig:framework}
\end{figure*}

\section{Method}
The key idea of our proposed framework is to learn an anonymization function such that it deteriorates the privacy attributes without requiring any privacy labels in the training, and maintains the performance of action recognition task. We build our self-supervised framework upon the existing supervised adversarial training framework of~\cite{wu_tpami}. A schematic of our framework is depicted in Fig.~\ref{fig:framework}. In Sec~\ref{sec:formulation}, we first formulate the problem by explaining our objective. In Sec~\ref{sec:framework} we present details of each component of our framework, and in Sec~\ref{sec:minimax} we explain the optimization algorithm employed in our framework.

\subsection{Problem Formulation}
\label{sec:formulation}
Let's consider a video dataset $X$ with action recognition as an utility task, $T$, and privacy attribute classification as a budget task, $B$. The goal of a privacy preserving action recognition system is to maintain performance of $T$, while cutting the budget $B$. This goal is achieved by learning an anonymization function, $f_{A}$, which transforms (anonymize) the original raw data $X$. Assume that the final system has \textit{any} action classification target model $f'_{T}$ and \textit{any} privacy target model $f'_{B}$. 
The goal of a privacy preserving training is to find an optimal point of $f_{A}$ called $f^{*}_{A}$, which is achieved by the following two criteria:

\noindent \textbf{C1:} $f^{*}_{A}$ should minimally affect the cost function of target model, $f'_{T}$, %
on raw data i.e. 
\begin{equation}
    L_{T}(f'_{T}(f^{*}_{A}(X)), Y_{T}) \approx L_{T}(f'_{T}(X), Y_{T}),
    \label{eq:ft_criterion}
\end{equation}
where $T$ denotes utility Task, $L_{T}$ is the loss function which is the standard cross entropy in case of single action label $Y_{T}$ or binary cross entropy in case of multi-label actions $Y_{T}$.

\noindent \textbf{C2:} Cost of privacy target model, $f'_{B}$, should increase on the transformed (anonymized) data compared to raw data i.e.
\begin{equation}
    L_{B}(f'_{B}(f^{*}_{A}(X)))  \gg  L_{B}(f'_{B}(X)),
    \label{eq:fb_criterion}
\end{equation}
where $B$ denotes privacy Budget, $L_{B}$ is the self-supervised loss for our framework, and binary cross entropy in case of a supervised framework, which requires privacy label annotations $Y_{B}$.

Increasing a self-supervised loss $L_{B}$ results in deteriorating \textit{all} useful information regardless of if it is about privacy attributes or not. However, the useful information for the action recognition is preserved via criterion \textbf{C1}. Combining criteria \textbf{C1} and \textbf{C2}, we can mathematically write the privacy preserving optimization equation as follows, where negative sign before $L_{B}$ indicates it is optimized by maximizing it:
\begin{equation}
    f^{*}_{A} = \operatorname*{argmin}_{f_{A}}[L_{T}(f'_{T}(f_{A}(X)), Y_{T}) -  L_{B}(f'_{B}(f_{A}(X)))].
    \label{eq:overall}
\end{equation}

\vspace{-5mm}
\subsection{Proposed Framework}
\label{sec:framework}
The proposed framework mainly consists of three components as shown in Fig~\ref{fig:framework}: (1) Anonymization function ($f_{A}$); (2) Self-supervised privacy removal branch; and (3) Action recognition or utility branch.
\vspace{-4mm}
\subsubsection{Anonymization Function ($f_{A}$)}
\vspace{-2mm}
The anonymization function is a learnable transformation function, which transforms the video in such a way that the transformed information can be useful to learn action classification on \textit{any} target model, $f'_{T}$, and not useful to learn \textit{any} privacy target model, $f'_{B}$. We utilize an encoder-decoder neural network as the anonymization function. $f_{A}$ is initialized as an identity function by training it using $\mathcal{L}_{L1}$ reconstruction loss as given below:
\begin{equation}
    \mathcal{L}_{L1} = \sum_{c=1}^{C}\sum_{h=1}^{H}\sum_{w=1}^{W}|x_{c,h,w}- \hat{x}_{c,h,w}|,
    \label{eq:l1loss}
\end{equation}
where, $x$ is input image, $\hat{x}$ is sigmoid output of $f_{A}$ logits, $C$ = input channels, $H$ = input height, and $W$ = input width.

\vspace{-4mm}
\subsubsection{Self-supervised privacy removal branch}
\begin{figure*}[h]
\begin{center}
  \includegraphics[width=0.80\textwidth]{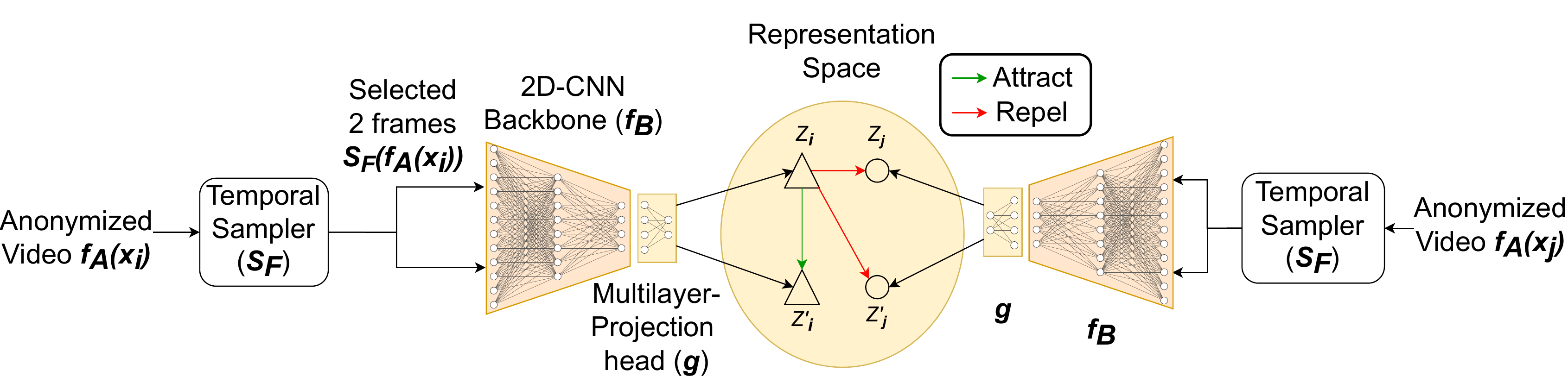}
\end{center}
\vspace{-0.3cm}
\caption{\textbf{Contrastive Self-supervised Loss} is used to maximize the agreement between two frames of a video and maximize disagreement between frames of different videos. Please refer to Sec~\ref{sec:sslbranch} for more details.}
\vspace{-3mm}
\label{fig:sslloss}
\end{figure*}

\vspace{-2mm}
\label{sec:sslbranch}
A schematic of self-supervised privacy removal branch is shown in Fig.~\ref{fig:sslloss}. First the video $x_{i}$ is passed through $f_{A}$ to get the anonymized video $f_{A}(x_{i})$, which is further passed through a temporal Frame sampler $S_{F}$.~$S_{F}$ samples 2 frames out of the video with various $S_{F}$ strategies, which are studied in Section~\ref{sec:ablation}. The sampled pair of frames ($S_{F}$($f_{A}$($x_{i}$))) are projected into the representation space through 2D-CNN backbone $f_{B}$ and a non-linear projection head $g(\cdot)$. The pair of frames of video $x_{i}$ corresponds to projection $Z_{i}$ and $Z'_{i}$ in the representation space. The goal of the contrastive loss is to maximize the agreement between projection pair ($Z_{i}$, $Z'_{i}$) of the same video $x_{i}$, while maximizing the disagreement between projection pairs of different videos ($Z_{i}$, $Z_{j}$), where $j \neq i$. The NT-Xent contrastive loss~\cite{simclr} for a batch of $N$ videos is given as follows:
\vspace{-2mm}
\begin{equation}\label{eq:ntxent}
  L_{B}^{i}=-\log \frac{h\left(Z_{i}, Z'_{i}\right)}{\sum_{j=1}^{N}[\mathbb{1}_{[j\neq i]} h(Z_{i}, Z_{j}) + h(Z_{i}, Z'_{j})]},
\end{equation}
\noindent where $h(u, v)=\exp \left(u^{T}v/(\|u\| \|v\| \tau) \right)$ is used to compute the similarity between $u$ and $v$ vectors with an adjustable parameter temperature, $\tau$. $\mathbb{1}_{[j\neq i]} \in \{0, 1\}$ is an indicator function which equals 1 iff $j \neq i$.

\vspace{-2mm}

\subsection{Minimax optimization}
\label{sec:minimax}
\vspace{-2mm}
In order to optimize the proposed self-supervised framework with the objective of Eq.~\ref{eq:overall}, let's consider anonymization function $f_{A}$ parameterized by $\theta_{A}$, and auxiliary models $f_{B}$ and $f_{T}$ respectively parameterized by $\theta_{B}$ and $\theta_{T}$. Assume, $\alpha_A$, $\alpha_B$, $\alpha_T$ respectively be the learning rates for $\theta_A$, $\theta_B$, $\theta_T$. First of all, $\theta_{A}$ is initialized as given below ( Eq.~\ref{eq:fa_init}), unless $f_{A}$ reaches to threshold $th_{A0}$ reconstruction performance ( Eq.~\ref{eq:l1loss}) on validation set:
\vspace{-2mm}
\begin{equation}
    \theta_A \gets \theta_A - \alpha_A \nabla_{\theta_A} (\mathcal{L}_{L1}(\theta_A)).
    \label{eq:fa_init}
    \vspace{-2mm}
\end{equation}
Once $\theta_{A}$ is initialized, it is utilized for initialization of $\theta_{T}$ and $\theta_{B}$ as shown in the following equations unless their performance reaches to the loss values of $th_{B0}$ and $th_{T0}$ :
\begin{equation}
    \theta_T \gets \theta_T - \alpha_T \nabla_{\theta_T} (L_{T}(\theta_T, \theta_A)),
    \label{eq:ft_init}
\end{equation}
\vspace{-5mm}
\begin{equation}
    \theta_B \gets \theta_B - \alpha_B \nabla_{\theta_B} (L_{B}(\theta_B, \theta_A)).
    \label{eq:fb_init}
\end{equation}
After the initialization, two step iterative optimization process takes place. The first step is depicted in the left side of Fig.~\ref{fig:framework}, where $\theta_{A}$ is updated using the following equation:
\vspace{-1mm}
\begin{equation}
    \theta_A \gets \theta_A - \alpha_A \nabla_{\theta_A} (L_T(\theta_A,\theta_T) - 	\omega L_B(\theta_A, \theta_B)),
    \label{eq:faupdate}
\end{equation}

\noindent where $\omega\in(0,1)$ is the relative weight of SSL loss, $L_{B}$, with respect to supervised action classification loss, $L_{T}$. Here the negative sign before $L_{B}$ indicates that we want to maximize it. In implementation, it can be simply achieved by using negative gradients~\cite{domainadaptation}.

In the second step, as shown in the right part of the Fig.~\ref{fig:framework}, $\theta_{T}$ and $\theta_{B}$ are updated using Eq.~\ref{eq:ft_init} and ~\ref{eq:fb_init}, respectively. We update $\theta_{B}$ to get powerful negative gradients in the next iteration's step-1. Note that there is a similarity with GAN training here; we can think of $f_{A}$ as the a generator who tries to fool $f_{B}$ in the first step and, in the second step $f_{B}$ tries to get stronger through update of Eq.~\ref{eq:fb_init}. This two step iterative optimization process continues until $L_{B}$ reaches to a maximum value $th_{Bmax}$.

\subsection{Intuition: SSL Branch and Privacy removal}
\vspace{-2mm}
Take a model $f_b$ \textit{initialized} with self-supervised contrastive loss (SSL) \textit{pretraining}. 
Now keeping $f_b$ {\em frozen}, when we try to {\em maximize} the contrastive loss, 
it changes the {\em input} to $f_b$ in such a way that it decreases \textit{agreement between frames} of the same video. \textit{We know that frames of the same video share a lot of semantic information}, and minimizing the agreement between them results in destroying (i.e. unlearning) most of the semantic info of the input video. In simple terms, maximizing contrastive loss results in destroying all highlighted \textit{attention map} parts of \supp{Supp. Fig 7, 8} middle row. Since this unlearned \textit{generic semantic info} contained privacy attributes related to human, scene, and objects; we end up removing \textit{private info} in the input. In this process, we also ensure that semantics related to action reco remains in video, through the action reco branch.

\vspace{-2mm}
\section{Training and Evaluation Protocols}
\vspace{-2mm}
The existing training and evaluation protocols are discussed in Sec~\ref{sec:knownprotocols}, ~\ref{sec:knowncrossdomain} and a new proposed generalization protocol is introduced in Sec~\ref{sec:novelprotocols}.

\vspace{-2mm}
\subsection{Same-dataset training and evaluation protocol}
\vspace{-2mm}
\label{sec:knownprotocols}

Training of supervised privacy preserving action recognition method requires a video dataset $X^{t}$ with action labels $Y^{t}_{T}$, and privacy labels $Y^{t}_{B}$, where $t$ denotes training set. Since, our self-supervised privacy removal framework does not require any privacy labels, we do not utilize $Y^{t}_{B}$. Once the training is finished, the anonymization function is now frozen, called $f^{*}_{A}$, and auxiliary models $f_{T}$ and $f_{B}$ are discarded. To evaluate the quality of the learned anonymization, $f^{*}_{A}$ is utilized to train: (1) a new action classifier $f'_{T}$ over the train set ($f^{*}_{A}(X^{t}), Y^{t}_{T}$); and (2) a new privacy classifier $f'_{B}$ to train over ($f^{*}_{A}(X^{t}), Y^{t}_{B}$). For clarification, we do not utilize privacy labels for training $f_{A}$ in any protocol. Privacy labels are used only for the evaluation purpose to train the target model $f'_{B}$. Once the target models $f'_{T}$ and $f'_{B}$ finish training on the anonymized version of train set, they are evaluated on test set ($f^{*}_{A}(X^{e}), Y^{e}_{T}$) and ($f^{*}_{A}(X^{e}), Y^{e}_{B}$), respectively, where $e$ denotes evaluation/test set. Test set performance of the action classifier is denoted as $A^{1}_{T}$ (Top-1 accuracy) or $A^{2}_{T}$ (classwise-mAP), and the performance of privacy classifier is denoted as $A^{1}_{B}$ (classwise-mAP) or $A^{2}_{B}$ (classwise-$F_{1}$). Detailed figures explaining different training and evaluation protocols are provided in~\supp{Supp.Sec.G}.  

\subsection{Cross-dataset training and evaluation protocol} 
\label{sec:knowncrossdomain}
\vspace{-2mm}
In practice, a trainable-scale video dataset with action and privacy labels doesn't exist. The authors of~\cite{wu_tpami} remedy the supervised training process by a cross-dataset training and/or evaluating protocol.  Two different datasets were utilized in ~\cite{wu_tpami}: action annotated dataset ($X^{t}_{action}, Y^{t}_{T}$) to optimize $f_{A}$ and $f_{T}$; and privacy annotated dataset ($X^{t}_{privacy}, Y^{t}_{B}$) to optimize $f_{A}$ and $f_{B}$. Again, note that in this protocol, our self-supervised framework does not utilize $Y^{t}_{B}$. After learning the $f_{A}$ through the different train sets, it is frozen and we call it $f^{*}_{A}$. A new action classifier $f'_{T}$ is trained on anonymized version of action annotated dataset ($f^{*}_{A}(X^{t}_{action}), Y^{t}_{T}$), and a new privacy classifier $f'_{B}$ is trained on the anonymized version of the privacy annotated dataset ($f^{*}_{A}(X^{t}_{privacy}), Y^{t}_{B}$). Once the target models $f'_{T}$ and $f'_{B}$ finish training on the anonymized version of train sets, they are respectively evaluated on test sets ($f^{*}_{A}(X^{e}_{action}), Y^{e}_{T}$) and ($f^{*}_{A}(X^{e}_{privacy}), Y^{e}_{B}$).

\subsection{Novel action and privacy attributes protocol}
\vspace{-2mm}
\label{sec:novelprotocols}
For the prior two protocols discussed above, the same training set $X^{t}$ ($X^{t}_{action}$ and $X^{t}_{privacy}$) is used for the target models $f'_{T}$, $f'_{B}$ and learning the anonymization function $f_{A}$. However, a learned anonymization function $f^{*}_{A}$ is expected to generalize on any action or privacy attributes. To evaluate the generalization on novel actions, an anonymized verion of novel action set $f^{*}_{A}(X^{nt}_{action})$, such that $Y^{nt}_{T}\cap Y^{t}_{T}= \phi$, is used to train the target action model $f'_{T}$, and its performance is measured on the anonymized test set of novel action set $f^{*}_{A}(X^{ne}_{action})$. For privacy generalization evaluation, a novel privacy set $f^{*}_{A}(X^{nt}_{privacy})$ (s.t. $Y^{nt}_{B}\cap Y^{t}_{B}= \phi$) (where $nt$ represents novel training) is used to train the privacy target model $f'_{B}$, and its performance is measured on novel privacy test set $f^{*}_{A}(X^{ne}_{privacy})$ (where $ne$. represents novel evaluation) Please note that novel privacy attribute protocol may not be referred as a \textit{transfer protocol} for the methods, which do not use privacy attributes $Y^{t}_{B}$ in learning $f_{A}$.

\section{Experiments}

\begin{table*}
\centering
\small
\begingroup
\setlength{\tabcolsep}{4pt} %
\renewcommand{\arraystretch}{1.0} %
\begin{tabular}{lccc|ccc|ccc} 
\hline

\hline

\hline\\[-3mm]
\multicolumn{1}{l}{\multirow{2}{*}{Method}} & \textbf{UCF101} & \multicolumn{2}{c|}{\textbf{VISPR1}} & \multicolumn{3}{c|}{\textbf{PA-HMDB}} & \multicolumn{3}{c}{\textbf{P-HVU}}  \\
\multicolumn{1}{c}{}                        &          Action               & \multicolumn{2}{c|}{Privacy}                        & Action & \multicolumn{2}{c|}{Privacy}                         & Action & Objects & Scenes  \\ 
                                            & Top-1 ($\uparrow$)                      & cMAP ($\downarrow$)   & F1 ($\downarrow$)                                 & Top-1 ($\uparrow$)    & cMAP ($\downarrow$)   & F1 ($\downarrow$)                          & cMAP ($\uparrow$)     & cMAP ($\downarrow$)     & cMAP ($\downarrow$)     \\
\hline
                                            
Raw data                                    & 62.33                    & 64.41  & 0.555                              & 43.6  & 70.1  & 0.401                         & 20.1   & 11.90    & 25.8    \\
Downsample-$2\times$                                       & 54.11                   & 57.23 & 0.483                               & 36.1  & 61.2  & 0.111                          & 10.9   & 2.45    & 8.6     \\
Downsample-$4\times$                                       & 39.65                  & 50.07 & 0.379                               & 25.8  & 41.4  & 0.081                          & 0.78   & 0.89    & 1.76    \\
Obf-Blackening                               & 53.13                   & 56.39 & 0.457                                & 34.2  & 63.8  & 0.386                         & 8.6    & 6.12    & 22.1    \\
Obf-StrongBlur                               & 55.59                    & 55.94 & 0.456                                & 36.4  & 64.4  & 0.243                         & 11.3   & 6.89    & 22.8    \\
Obf-WeakBlur                               & 61.52                    & 63.52  & 0.523                                & 41.7  & 69.4  & 0.398                         & 18.6   & 11.33   & 25.4    \\
Noise-Features~\cite{icdm19}                              & 61.90                    & 62.40  & 0.531                                & 41.5  & 69.1  & 0.384                         & --   & --   & --    \\
Supervised~\cite{wu_tpami}                                    & 62.10                    & 55.32\increase{\textbf{14\%}} & 0.461\increase{\textbf{17\%}}                              & 42.3  & 62.3\increase{11\%} & 0.194\increase{51\%}                          & 18.33  & 1.98\increase{83\%}    & 9.5\increase{\textbf{63\%}}    \\

{\bf Ours }                                      & 62.03                   & 57.43\increase{11\%}  & 0.473\increase{15\%}                               & 43.1 & 62.7\increase{11\%} & 0.176\increase{\textbf{56\%}}                         & 18.01  & 1.42\increase{\textbf{88\%}}    & 9.91\increase{62\%}    \\
\hline

\hline

\hline\\[-3mm]
\end{tabular}
\endgroup
\vspace{-3mm}
\caption{Comparison of existing privacy preserving action recognition method on \textbf{known action and privacy attributes} protocol. Our framework achieves a competitive performance to the supervised method~\cite{wu_tpami}. \increase{\%} denotes relative drop from raw data. For a graphical view, refer to \supp{Supp.Sec.D}.}
\label{table:samedomain}
\end{table*}

\begin{table*}
\centering
\small
\begin{tabular}{lcccc|cc} 
\hline

\hline

\hline\\[-3mm]
\multirow{3}{*}{Method} & \multicolumn{2}{c}{Transfer Evaluation: \textbf{Action}} & \multicolumn{2}{c|}{Transfer Evaluation: \textbf{Privacy}} & \multicolumn{2}{c}{Transfer Evaluation P-HVU} \\
                        & \textbf{UCF$\rightarrow$HMDB} & \textbf{UCF$\rightarrow$PA-HMDB }                      & \multicolumn{2}{c|}{\textbf{VISPR1$\rightarrow$VISPR2}}               & Action & \textbf{Scenes $\rightarrow$ Obj }                         \\
                        & Top-1(\%) ($\uparrow$)        & Top-1(\%) ($\uparrow$)                                  & cMAP(\%) ($\downarrow$)  & F1 ($\downarrow$)                                       & cMAP(\%) ($\uparrow$)     & cMAP(\%) ($\downarrow$)                                   \\ 
\hline
Raw data                & 35.6      & 43.6                                & 57.6 & 0.498                                      & 20.1   & 11.9                                  \\
Downsample-$2\times$                   & 24.1      & 36.1                                & 52.2 & 0.447                                      & 10.9   & 2.45                                  \\
Downsample-$4\times$                   & 16.8      & 25.8                                & 41.5 & 0.331                                      & 0.78   & 0.89                                  \\
Obf-Blackening          & 26.2      & 34.2                                & 53.6 & 0.46                                       & 8.6    & 6.12                                  \\
Obf-StrongBlur           & 26.4      & 36.4                                & 53.7 & 0.462                                      & 11.3   & 6.89                                  \\
Obf-WeakBlur           & 33.7      & 41.7                                & 55.8 & 0.486                                      & 18.6   & 11.33                                 \\
Noisy Features~\cite{icdm19}                & 31.2      & 41.5                                & 53.7 &  0.458                                      &  --  & --                                  \\
Supervised~\cite{wu_tpami}                & 33.2      & 40.6                                & 49.6\increase{14\%} & 0.399\increase{20\%}                                      & 18.34  & 6.43\increase{46\%}                                  \\
{\bf Ours}                    & 34.1     & 42.8                                & \textbf{47.1}\increase{\textbf{18\%}} & \textbf{0.386}\increase{\textbf{22\%}}                                      & 18.01  & \textbf{1.42}\increase{\textbf{88\%}}                                  \\
\hline

\hline

\hline\\[-3mm]
\end{tabular}
\vspace{-3mm}

\caption{Comparison of existing privacy preserving action recognition method on \textbf{novel action and privacy attributes} protocol. Our framework outperforms the supervised method~\cite{wu_tpami}. \increase{\%} denotes relative drop from raw data.}
\label{table:diffdomain}
\vspace{-1.5em}
\end{table*}

\subsection{Datasets}
\vspace{-2mm}
\noindent\textbf{UCF101}~\cite{ucf101} and \textbf{HMDB51}~\cite{hmdb} are two of the most commonly used datasets for the human action recognition.

\noindent \textbf{PA-HMDB}~\cite{wu_tpami} is dataset of 515 videos annotated with video level action annotation and framewise human privacy annotations. The dataset consists of 51 different actions and 5 different human privacy attributes.

\noindent \textbf{P-HVU} is a selected subset of LSHVU~\cite{hvu}, which is a large-scale dataset of multi-label human actions, with a diverse set of auxiliary annotations provided for object, scenes, concepts, event etc. However, the all auxiliary annotations are not provided for all videos. We select a subset of action-object-scene labels based on their availability in the val set to create our train/test split. The dataset consists of 739 action classes, 1678 objects, and 248 scene categories. Train/test split of P-HVU consists of 245,212/16,012 videos to provide a robust evaluation.

\noindent \textbf{VISPR}~\cite{vispr} is an image dataset with a diverse set of personal information in an image like skin color, face, gender, clothing, document information etc.

\noindent Further details are provided in \supp{Supp.Sec.B}.

\subsection{Implementation Details}
\vspace{-2mm}
For default experiment setting, we utilize UNet~\cite{unet} as $f_{A}$, R3D-18~\cite{kenshohara} as $f_{T}$, and ResNet-50~\cite{resnet} as $f_{B}$. 
For a fair evaluation we report results of different methods with the exact same training augmentations and model architectures. Implementation details related to training setting, hyperparmeters, and model architectures can be found in \supp{Supp.Sec.C}. Visualization of the learned anonymization from different methods can be seen in \supp{Supp.Sec.F}.  

\noindent \textbf{Downsampling methods} are adopted with a down-sampled versions of input resolution with a factor of $2\times$ and $4\times$ used in training and testing.

\noindent \textbf{Obfuscation methods} are carried out using a MS-COCO~\cite{coco} pretrained Yolo~\cite{yolo} object detector to detect person category. The detected persons are removed using two different obfuscation strategies: (1) blackening the detected bounding boxes; (2) applying Gaussian blur in the detected bounding boxes at two different strengths.

\subsection{Evaluating learned anonymization on known action and privacy attributes}
\vspace{-2mm}
For known action and privacy attributes, we follow Section~\ref{sec:knowncrossdomain} to evaluate on 2 existing protocols from~\cite{wu_tpami}, and follow Section~\ref{sec:knownprotocols} to present a new protocol using P-HVU dataset for same-domain training and testing. Results are shown in Table~\ref{table:samedomain}.

\noindent\textbf{UCF101-VISPR cross dataset training and evaluation} 

\noindent In this protocol, $X^{t}_{action}$= UCF101 trainset and $X^{e}_{action}$= UCF101 testset; $X^{t}_{privacy}$ = VISPR trainset and
$X^{e}_{privacy}$= VISPR testset.\\
\vspace{-3mm}

\noindent \textbf{HMDB51-VISPR cross dataset training and PA-HMDB evaluation} 
\noindent In this protocol, $X^{t}_{action}$ = HMDB51 trainset, and $X^{e}_{action}$ = PA-HMDB, $X^{t}_{privacy}$ = VISPR trainset, and 
$X^{e}_{privacy}$ = PA-HMDB. \\

\vspace{-3mm}
\noindent \textbf{P-HVU same dataset training and evaluation}

\noindent In this protocol, utility task is multi-label action recognition and privacy is defined in terms of object and scene multi-label classification. In this protocol, $X^{t}$ = P-HVU trainset, and $X^{e}$ = P-HVU testset.

We can observe in Table~\ref{table:samedomain} that our proposed {\em self-supervised} framework achieves a comparable action-privacy trade-off in case of known action and privacy attributes. Other methods like \texttt{Downsample-$4\times$}, \texttt{Obf-blackening} and \texttt{Obf-StrongBlur} get a commendable privacy removal, however, at a cost of action recognition performance.

\subsection{Evaluating learned anonymization on Novel action and privacy attributes}
\vspace{-2mm}
Following Sec.~\ref{sec:novelprotocols}, we propose 2 protocols for the novel actions and 2 protocols for the novel privacy attributes. 

\begin{figure}[h]
\vspace{-2em}
\begin{center}
\includegraphics[width=0.9\columnwidth]{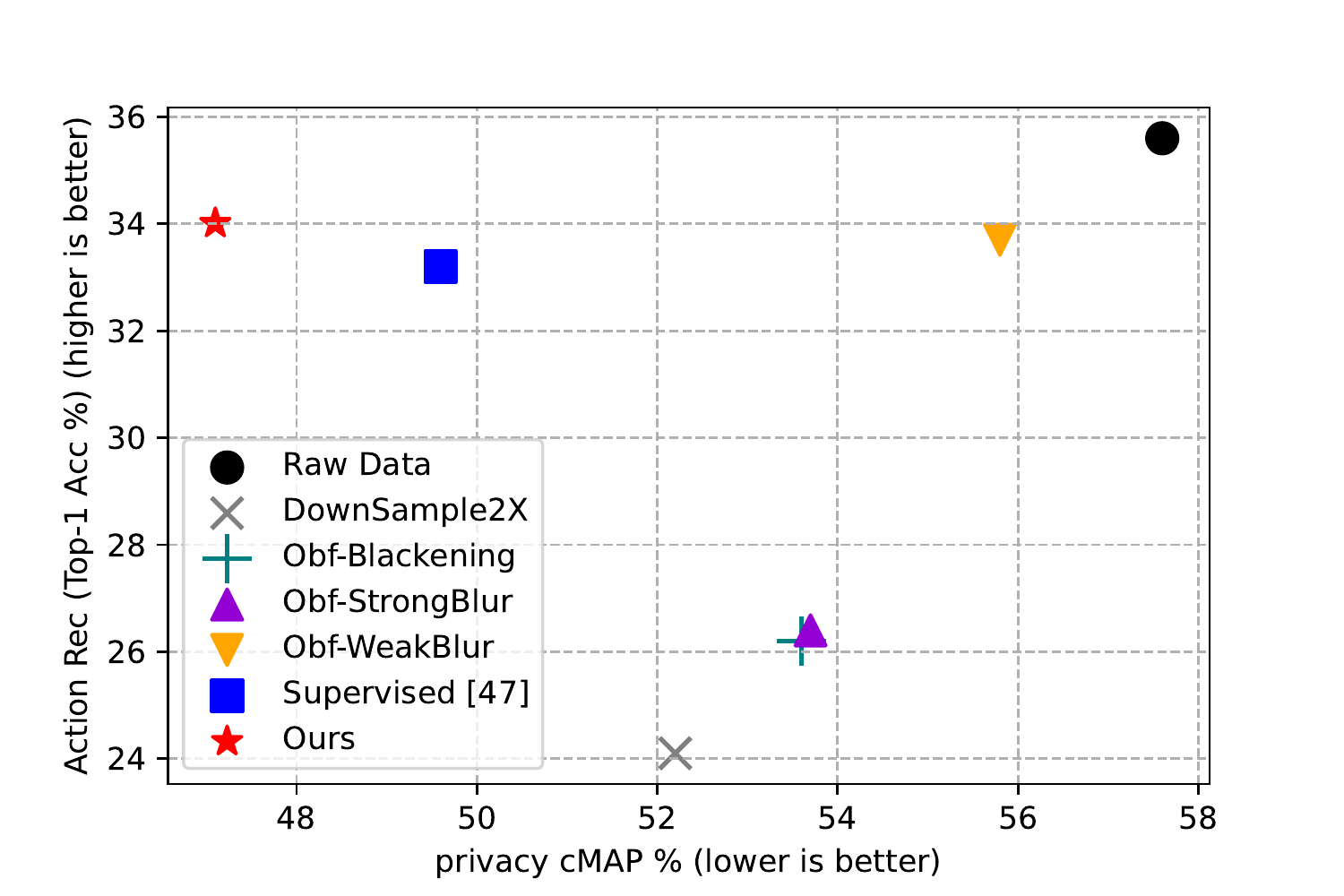}

\end{center}
\vspace{-0.6cm}

\caption{Trade-off between action classification and privacy removal while generalizing from UCF101$\rightarrow$HMDB51 for action and VISPR1$\rightarrow$VISPR2 for privacy attributes. Our self-supervised method achieves the best trade-off among other methods.}

\label{fig:tradeoffplot}
\vspace{-5mm}

\end{figure}

\noindent \textbf{Novel action and privacy attributes} In this protocol, for actions $X^{t}_{action}$ = UCF101 trainset, $X^{nt}_{action}$ = HMDB51 trainset, $X^{ne}_{action}$ = HMDB51 testset/ PA-HMDB and for privacy, $X^{t}_{privacy}$ = VISPR-1 trainset, $X^{nt}_{privacy}$ = VISPR-2 trainset and $X^{ne}_{privacy}$ = VISPR-2 testset. From the left part of Table~\ref{table:diffdomain} and Fig.~\ref{fig:tradeoffplot}, we can observe that our method outperforms the supervised method~\cite{wu_tpami} in both action and privacy attribute generalization. 

\noindent \textbf{Novel privacy attributes from Scenes to Objects}
\noindent In this protocol, we take known action set $X^{t}_{action}$ = P-HVU trainset, and $X^{e}_{action}$ = P-HVU testset, $X^{t}_{privacy}$ = P-HVU trainset Object, $X^{nt}_{privacy}$ = P-HVU trainset Scene and $X^{ne}_{privacy}$ = P-HVU testset Scene. We can observe from the right most part of Table~\ref{table:diffdomain} that while testing the learned anononymization from scenes to objects, supervised method ~\cite{wu_tpami} gets a similar results like Obf-StrongBlur and removes only \textbf{$\sim$46\%} of the raw data's privacy, whereas our method removes \textbf{$\sim$88\%} object privacy of the raw data. Main reason for difference in our method's performance gain over~\cite{wu_tpami} in Table~\ref{table:diffdomain} is due to the {\em amount of domain shift} in \textit{novel} privacy attributes. In VISPR1$\rightarrow$2, domain shift is very small eg SkinColor(V1)$\rightarrow$Tattoo(V2) (\supp{Supp.Table 1}), and hence [37] is still able to generalize and perform only ($<$5\%) worse than our method. Whereas, in PHVU Scene$\rightarrow$Obj, domain shift is huge eg TennisCourt (Scene)$\rightarrow$TennisRacket (Obj), where [37] suffers in generalizing and performs significantly ($>$40\%) poor than ours.
Additional experiments can be found in \supp{Supp.Sec.D} and qualitative results can be found in \supp{Supp.Sec.F}.

\subsection{Ablation Study}
\vspace{-1mm}

\label{sec:ablation}
\noindent \textbf{Experiments with different privacy removal branches}
Second row in Table~\ref{table:ablFramework} %
shows the results just using an encoder-decoder based model $f_{A}$ without using any privacy removal branch $f_{B}$. 
However, the style changing fails to anonymize privacy information. %
In our next attempt, we utilize a pretrained SSL \textit{frozen} model to anonymize the privacy information by Eq.~\ref{eq:faupdate}. This method of frozen $f_{B}$ is able to remove the privacy information by a small extent ($<2\%$), however, our biggest boost in privacy removal ($\mathbf{7\%}$) comes from updating $f_{B}$ with every update in $f_{A}$ as can be seen in the second last row of Table~\ref{table:ablFramework}. This observation shows the importance of updating the $f_{B}$ in step-2 (Eq.~\ref{eq:fb_init}) of minimax optimization. In other words, we can say that if $f_{B}$ is not updated with $f_{A}$, then it becomes very easy for $f_{A}$ to fool $f_{B}$ i.e. maximize $L_{B}$, which ultimately leads to a poor privacy removal. Additionally, we also experiment with a spatio-temporal SSL framework as privacy removal branch. Details are given in \supp{Supp.Sec.C}. Note that removing spatio-temporal semantics from the input video leads to severe degradation in action recognition performance, which is the main reason of choosing 2D SSL privacy removal branch in our framework in order to remove only spatial semantics from the input video. 
\vspace{-2mm}

\begin{table}[h]
\centering
\small
\begin{tabular}{llccc} 
\hline

\hline

\hline\\[-3mm]
$f_{A}$ & $f_{B}$ & \textbf{UCF101} & \multicolumn{2}{c}{\textbf{VISPR1}}  \\
        &         & Top-1 ($\uparrow$)     & cMAP ($\downarrow$)   & F1 ($\downarrow$)                \\ 
\hline
\xmark & \xmark                                    & 62.3   & 64.4  & 0.555             \\
\cmark & \xmark                    & 63.5   & 64.1  & 0.549               \\
\cmark & Spatial (Frozen)                                     & 62.2   & 62.2  & 0.535               \\
\cmark & \textbf{Spatial}                                  & \textbf{62.1}  & \textbf{57.4} & \textbf{0.473}              \\
\cmark & Spatio-Temporal                                  & 56.4  & 56.6 & 0.461              \\
\hline

\hline

\hline\\[-3mm]
\end{tabular}
\vspace{-3mm}
\caption{Experiments with different privacy removal branches}
\label{table:ablFramework}
~\vspace{-5mm}
\end{table}

\noindent \textbf{Temporal sampling strategies for SSL}
In order to experiment with various Temporal sampler ($S_{F}$) for choosing a pair of frames from a video, we change the duration (distance) between the two frames as shown in Table~\ref{table:framesampling}. The chosen pair of frames from a video is considered for the positive term of contrastive loss (Eq.~\ref{eq:ntxent}). In our default setting of experiments, we randomly select a pair of frames from a video as shown in the first row. We observe that mining positive frames from further distance decreases the anonymization capability. This is because mining the very dissimilar positives in contrastive loss leads to poorly learned representation, which is also observed while taking temporally distant positive pair in ~\cite{cvrl, Feichtenhofer_2021_CVPR}.  
\begin{table}[h]
\centering
\small
\begin{tabular}{lccc} 
\hline

\hline

\hline\\[-3mm]
\multirow{2}{*}{\begin{tabular}[c]{@{}l@{}}Distance between \\positive frames\end{tabular}} & \textbf{UCF101} & \multicolumn{2}{c}{\textbf{VISPR1}}  \\
                                                                                            & Top-1(\%) ($\uparrow$)     & cMAP(\%) ($\downarrow$)   & F1 ($\downarrow$)                \\ 
\hline
No constraint                                                                               & 62.1  & 57.4 & 0.473              \\
$>$64 frames                                                                                   & 62.1   & 58.7  & 0.488               \\
$<$8 frames                                                                                    & 63.4   & 57.1  & 0.443               \\
\hline

\hline

\hline\\[-3mm]
\end{tabular}
\vspace{-3mm}
\caption{Effect of frame sampling strategy in contrastive loss of SSL privacy removal branch}
\label{table:framesampling}
~\vspace{-5mm}
\end{table}

\noindent \textbf{Effect of different SSL frameworks}
As shown in Table~\ref{table:sslframework}, we experiment with three different 2D SSL schemes in Eq.~\ref{eq:ntxent}. We can observe that NT-Xent~\cite{simclr} and MoCo~\cite{moco} achieve comparable performances, however, RotNet~\cite{rotnet} framework provides a suboptimal performance in both utility and privacy. Our conjecture is that this is because RotNet mainly encourages learning global representation, and heavily removing the global information from the input via privacy removal branch leads to drop in action recognition performance as well. 
\begin{table}[h]
\centering
\small
\begin{tabular}{lccc} 
\hline

\hline

\hline\\[-3mm]
\multirow{2}{*}{\textbf{SSL Loss}} & \textbf{UCF101} & \multicolumn{2}{c}{\textbf{VISPR1}}  \\
                          & Top-1(\%) ($\uparrow$)     & cMAP(\%) ($\downarrow$)   & F1 ($\downarrow$)                \\ 
\hline
NT-Xent~\cite{simclr}                   & 62.1  & 57.4 & 0.473              \\
MoCo~\cite{moco}                      & 61.4   & 57.1  & 0.462               \\
RotNet~\cite{rotnet}                    & 58.1  & 60.2  & 0.504               \\
\hline

\hline

\hline\\[-3mm]
\end{tabular}
\vspace{-3mm}
\caption{Effect of different SSL frameworks}
\label{table:sslframework}
~\vspace{-4mm}
\end{table}

\noindent \textbf{Effect of different $f_{B}$ and $f_{T}$ architectures} To understand the effect of auxiliary model $f_{B}$ in the training process of $f_{A}$, we experiment with different privacy auxiliary models $f_{B}$, and report the performance of their learned $f^{*}_{A}$ in the same evaluation setting as shown in Table~\ref{table:fb}. We can observe that using a better architecture of $f_{B}$ leads to better anonoymization. There is no significant effect of using different architectures of $f_{T}$ in learning $f_{A}$ (\supp{Supp.Sec.E}).

\begin{table}[h]
\vspace{-0.5em}
\centering
\small
\begin{tabular}{lccc} 
\hline

\hline

\hline\\[-3mm]
\multirow{2}{*}{$f_{B}$ \textbf{architecture}} & \textbf{UCF101} & \multicolumn{2}{c}{\textbf{VISPR1}}  \\
                                 & Top-1 ($\uparrow$)     & cMAP ($\downarrow$)   & F1 ($\downarrow$)                \\ 
\hline
MobileNetV1 (MV1)                     & 62.1   & 58.14 & 0.488               \\
ResNet50 (R50)                        & 62.1  & 57.43 & 0.473              \\
R50 + MV1         & 61.4   & 56.20  & 0.454               \\
\hline

\hline

\hline\\[-3mm]
\end{tabular}
\vspace{-3mm}
\caption{Effect of different $f_{B}$ in minimax optimization}
\label{table:fb}
\end{table}

~\vspace{-10mm}
\section{Limitation}
\vspace{-2mm}
One limitation of our work is that it utilizes the basic frameworks for self-supervised learning, and which may be suitable only for the action recognition, and not directly suitable for other video understanding tasks like actions detection or action anticipation. Additionally, there is still room of improvement to match the supervised baseline in case of known action-privacy attributes. 

\section{Conclusion}
\vspace{-2mm}
We introduced a novel self-supervised privacy preserving action recognition framework which does not require privacy labels for the training. Our extensive experiments show that our framework achieves competitive performance compared to the supervised baseline for the known action-privacy attributes. We also showed that our method achieves better generalization to novel action-privacy attributes compared to the supervised baseline. Our paper underscores the benefits of contrastive self-supervised learning in privacy preserving action recognition.

\section*{Acknowledgments}
\vspace{-2mm}
We thank Vishesh Kumar Tanvar, Tushar Sangam, Rohit Gupta, and Zhenyu Wu for constructive suggestions.

\clearpage
{\small
\bibliographystyle{ieee_fullname}
\bibliography{main}
}
\clearpage

\appendix

\section{Supplementary Overview}

The supplementary material is organized into the following sections:

\begin{itemize}
    \item Section~\ref{sec:dataset}: Dataset details
    \item Section~\ref{sec:impl_details}: Implementation details such as network architectures, data augmentations, training setup, baseline implementation details, performance metrics. 
    \item Section~\ref{sec:addtional_results}: Evaluating learned anonymization function in various target settings. 

    \item Section~\ref{sec:addtional_ablation}: Additional ablation experiments.

    \item Section~\ref{sec:qualitative}: Qualitative results of the learned anonymization
    \item Section~\ref{sec:protocol}: Visual Aid to understand the training and evaluation protocol
    
\end{itemize}

\section{Datasets}
\label{sec:dataset}
\noindent\textbf{UCF101}~\cite{ucf101} has around 13,320 videos representing 101 different human activities. All results in this paper are  reported on split-1, which has 9,537 train videos and 3,783 test videos.

\noindent\textbf{HMDB51}~\cite{hmdb} is a relatively smaller action recognition dataset having 6,849 total videos collected from 51 different human actions. All results in this paper are reported on split-1, which has 3,570 train videos and 1,530 test videos.

\noindent\textbf{VISPR}~\cite{vispr} is an image dataset with a diverse set of personal information in an image like skin color, face, gender, clothing, document information etc. We use two subsets of privacy attributes of VISPR dataset as shown in Table~\ref{table:dataset}. Each of the privacy attribute is a binary label, where 0 indicates absence of the attribute and 1 indicates presence of the attribute in the image. An image can have multiple privacy attributes, hence it is as a multi-label classification problem. 
\begin{table}[h]
\centering
\begin{tabular}{ll} 
\hline

\hline

\hline\\[-3mm]
\textbf{VISPR1~\cite{wu_tpami}}             & \textbf{VISPR2}              \\ 
\hline
\texttt{a17\_color}         & \texttt{a6\_hair\_color}     \\
\texttt{a4\_gender}         & \texttt{a16\_race}           \\
\texttt{a9\_face\_complete} & \texttt{a59\_sports}         \\
\texttt{a10\_face\_partial} & \texttt{a1\_age\_approx}     \\
\texttt{a12\_semi\_nudity}  & \texttt{a2\_weight\_approx}  \\
\texttt{a64\_rel\_personal} & \texttt{a73\_landmark }      \\
\texttt{a65\_rel\_soci}   & \texttt{a11\_tattoo}         \\
\hline

\hline

\hline\\[-3mm]
\end{tabular}
\caption{Privacy attributes of VISPR~\cite{vispr} subsets.}
\label{table:dataset}
\end{table}

\noindent\textbf{PA-HMDB51}~\cite{wu_tpami} is subset of HMDB51 dataset with 51 action labels and 6 human privacy attributes which are annotated temporally. The privacy attributes are the same as VISPR-1 subset shown in Table~\ref{table:dataset} except \texttt{a65\_rel\_soci} attribute. Each privacy attribute has a fine-grained class assigned  as well, however, it is not considered in this paper. Following ~\cite{wu_tpami}, we use binary label for each privacy attribute i.e. if the privacy attribute is present in the image or not. 

\noindent \textbf{P-HVU} is a selected subset of LSHVU~\cite{hvu}, which is a large-scale dataset of multi-label human action with a diverse set of auxiliary annotations provided for objects, scenes, concepts, events etc. We consider using this dataset to understand privacy leakage in terms of object or scene. P-HVU is prepared from LSHVU dataset such that each video has object and scene annotations along with the action label. A video of the LSHVU always has action labels, however, it does not necessarily have scene and object label. We consider following steps to prepare P-HVU dataset:
\begin{itemize}
    \item Select all LSHVU {\em validation} set videos such that each video has object and scene annotation and call it \textit{P-HVU test set}. 
    \item Select LSHVU {\em train} set videos which has action, object and privacy class from the P-HVU test set, and filter out videos if either of the object or scene annotations are missing in the video and call it \textit{P-HVU train set}. 
\end{itemize}
Each video of the P-HVU dataset has multi-label action, object and scene annotation. The dataset consists of 739 action classes, 1678 objects, and 248 scene categories. Train/test split of P-HVU consists of 245,212/16,012 videos to provide a robust evaluation.

\section{Implementation Details}
\label{sec:impl_details}
\subsection{Architectural details}
For anonymization function we utilize PyTorch implementation\footnote{\href{https://github.com/milesial/Pytorch-UNet}{https://github.com/milesial/Pytorch-UNet}} of UNet~\cite{unet} with three output channels. For 2D-CNN based ResNet~\cite{resnet}, 3D-CNN models R3D-18~\cite{kenshohara}, and R2plus1D-18~\cite{r2plus1d}, we utilize \texttt{torchvision.models} implementation\footnote{\href{https://github.com/pytorch/vision/tree/main/torchvision/models}{https://github.com/pytorch/vision/tree/main/torchvision/models}}. Multi-layer projection head $\mathbf{g(\cdot)}$ of self-supervised privacy removal branch consists of 2 layers: Linear($2048$, $2048$) with ReLU activation and Linear($2048$, $128$) followed by L2-Normalization.

\subsection{Augmentations}
We apply two different sets of augmentation depending upon the loss function: (1) For supervised losses, we use standard augmentations like random crop, random scaling, horizontal flip and random gray-scale conversion with less strength. (2) For self-supervised loss, in addition to the standard augmentations with with more strength, we use: random color jitter, random cut-out and random color drop. For more details on augmentation strengths in supervised and self-supervised losses refer SimCLR~\cite{simclr}. In order to ensure temporal consistency in a clip, we apply the exact same augmentation on all frames of the clip. All video frames or images are resized to $112\times112$. Input videos are of 16 frames with skip rate of 2.

\subsection{Hyperparameters}

We use a base learning rate of 1e-3 with a learning rate scheduler which drops learning rate to its 1/10th value on the loss plateau. 

For self-supervised privacy removal branch, we use the 128-D output as representation vector to compute contrastive loss of temperature $\tau=0.1$. For RotNet~\cite{rotnet} experiment we use 4 rotations: \{0, 90, 180, 270\}.

\subsection{Training details}
To optimize parameters of different neural networks we use Adam optimizer~\cite{adam}. For initialization, we train $f_{A}$ for 100 epochs using $\mathcal{L}_{1}$ reconstruction loss, action recognition auxiliary model $f_{T}$ using cross-entropy loss for 150 epochs, and privacy auxiliary model $f_{B}$ using NT-Xent loss for 400 epochs. Training phase of anonymization function $f_{A}$ is carried out for 100 epochs, whereas target utility model $f'_{T}$ and target privacy model $f'_{B}$ are trained for 150 epochs. 

\subsection{Performance Metrics}
To evaluate the performance of target privacy model $f'_{B}$ we use macro-average of classwise mean average precision (cMAP). The results are also reported in average F1 score across privacy classes. F1 score for each class is computed at confidence 0.5. For action recognition, we use top-1 accuracy computed from video-level prediction from the model and groundtruth. A video-level prediction is average prediction of 10 equidistant clips from a video. 

\subsection{Baselines}
\noindent\textbf{Supervised adversarial framework~\cite{wu_tpami}:} we refer to official github repo\footnote{\href{https://github.com/VITA-Group/Privacy-AdversarialLearning}{https://github.com/VITA-Group/Privacy-AdversarialLearning}} and with the consultation of authors we reproduce their method. For fair comparison, we use exact same model architectures and training augmentations. For more details on hyperparameters refer~\cite{wu_tpami}. 

\noindent\textbf{Blurring based obfuscation baselines:} we first detect the person using MS-COCO~\cite{coco} pretrained yolov5x~\cite{yolov5} model in each frame of the video. After detecting the person bounding boxes, we apply Gaussian blur filter on the bounding boxes regions. We utilize \texttt{torchvision.transforms.GaussianBlur} function with kernel size = 21 and sigma = 10.0 for Strong blur, and kernel size = 13, sigma = 10.0 for the Weak blur baselines. For VISPR dataset, we first downsample images such that smaller side of image = 512. 

\noindent\textbf{Blackening based obfuscation baselines:} we first detect person bounding boxes using yolov5x model and assign zero value to all RGB channels of the bounding box regions. 

\noindent\textbf{Blackening based obfuscation baselines:} we first detect person bounding boxes using yolov5x model and assign zero value to all RGB channels of the bounding box regions. 
\noindent\textbf{Ablation with spatio-temporal privacy removal branch:} For ablation of \supp{Table 3} of the \supp{main paper}, we use naive extension of SimCLR~\cite{simclr} to the domain of video, where we consider two clips from the same video as positive and clips from other videos as negatives in the contrastive loss. R3D-18 is chosen as 3D-CNN backbone and MLP $\mathbf{g(\cdot)}$ consist of Linear($512$, $512$) with ReLU activation and Linear($512$, $128$) followed by L2-Normalization.

\noindent\textbf{Noisy Features baseline~\cite{icdm19}:}
 Zhang \etal~\cite{icdm19} proposed non-visual privacy preservation in wearable device from 1D singal of mobile sensors. We extended this work to video privacy by replacing LossNet to R3D-18, TransNet to UNet and extended similarity losses to handle video input.

\section{Additional results}
\label{sec:addtional_results}
\subsection{Evaluating $f^{*}_{A}$ privacy target model with $f'_{B}$ pretrained on a raw data }

\begin{table*}
\centering
\begin{tabular}{lcccccc} 
\hline

\hline

\hline\\[-3mm]
\multirow{2}{*}{Method} & \multicolumn{2}{c}{\textbf{VISPR1}} & \multicolumn{2}{c}{\textbf{VISPR2}} & \multicolumn{2}{c}{\textbf{PA-HMDB}}  \\
                        & cMAP (\%)($\downarrow$) & F1($\downarrow$)                      & cMAP (\%)($\downarrow$) & F1($\downarrow$)                      & cMAP (\%)($\downarrow$) & F1($\downarrow$)                        \\ 
\hline
Raw data                & 64.40     & 0.5553                  & 57.60     & 0.4980                  & 70.10     & 0.4010                    \\
Downsample-$2\times$                   & 51.23     & 0.4627                  & 46.39     & 0.4330                  & 60.04     & 0.2403                    \\
Downsample-$4\times$                   & 38.82     & 0.3633                  & 33.42     & 0.3055                  & 0.59      & 0.2630                    \\
Obf-Blackening          & 48.38     & 0.3493                  & 44.01     & 0.3134                  & 55.66     & 0.0642                    \\
Obf-StrongBlur           & 54.44     & 0.4440                  & 50.31     & 0.3990                  & 60.13     & 0.2830                    \\
Supervised~\cite{wu_tpami}                & 22.81\textbf{\increase{65\%}}     & 0.2437\increase{56\%}                  & 26.61\increase{54\%}     & 0.1840\increase{63\%}                  & 57.01\textbf{\increase{19\%}}     & 0.2310\increase{42\%}                    \\
{\bf Ours}                    & 27.44\increase{57\%}       & 0.0760\textbf{\increase{86\%} }                 & 20.02\textbf{\increase{65\%}}     & 0.0460\textbf{\increase{91\%}}                  & 58.90\increase{16\%}     & 0.0940 \textbf{\increase{77\%} }                  \\
\hline

\hline

\hline\\[-3mm]

\end{tabular}

\caption{Evaluating learned anonymization function $f^{*}_{A}$ to measure its privacy leakage from a \textbf{raw-data pretrained privacy target model $f'_{B}$}. Lower privacy classification score is better, \increase{\%} denotes relative drop from raw data. Our self-supervised gets a competitive performance to the supervised method~\cite{wu_tpami}.}
\label{table:fixed_fb}

\end{table*}

In a practical scenario, learned anonymization $f^{*}_{A}$ is not accessible to a intruder, hence one can try to extract privacy information using a pretrained privacy classifier of raw data. In this protocol, instead of learning a target privacy model $f'_{B}$ from the anonymized version of the training data, we directly evaluate $f^{*}_{A}$ using a privacy target model which is pretrained on raw data.  Results are shown in Table~\ref{table:fixed_fb}. We use ResNet-50 model as privacy target model, which is pretrained on raw training data of the the respective evaluation set. There are two main observations in in this protocol: (1) Compared to other methods, supervised~\cite{wu_tpami} and our self-supervised method gets a remarkable amount of privacy classification drop, which is desired to prevent privacy leakage. (2) Our method gets a competitive cMAP performance to ~\cite{wu_tpami}, and greatly outperforms it in terms of F1 score.

\subsection{Evaluating learned $f^{*}_{A}$ on different utility target model $f'_{T}$}
A learned anonymization function, $f^{*}_{A}$, should allow learning any action recognition target model, $f'_{T}$, over the anonymized version of training data without significant drop in the performance. Using the R3D-18 as a auxiliary action recognition model, $f_{T}$, in the training of anonymization function, we evaluate the learned $f^{*}_{A}$ to train different action recognition (utility) target models like R3D-18, C3D~\cite{c3d}, and R2plus1D-18 from scratch and  Kinetics-400~\cite{kinetics} pretraining. Results are shown in Table~\ref{table:f't}. We can observe that our method maintains the action recognition performance on any utility action recognition model. Also, it is interesting to notice that the learned anonymization by our method and method in ~\cite{wu_tpami} get benefit from a large-scale raw data pretraining of Kinetics-400.

\begin{table}
\centering
\small
\begingroup
\setlength{\tabcolsep}{2pt} %
\renewcommand{\arraystretch}{1.0}
\begin{tabular}{lcccc} 
\hline

\hline

\hline\\[-3mm]
Method     & \textbf{R3D-18} & \textbf{R2Plus1D} & \begin{tabular}[c]{@{}c@{}}\textbf{R2Plus1D}\\\textbf{K400 pretraining}\end{tabular} & \textbf{C3D}  \\ 
\hline
Raw data   & 62.3            & 64.33                & 88.76                                                                                   & 58.51                \\
Supervised~\cite{wu_tpami} & 62.1           & 62.58                & 85.33                                                                                   & 56.30                \\
\textbf{Ours}       & 62.03           & 62.71                & 85.14                                                                                   & 56.10                \\
\hline

\hline

\hline\\[-3mm]
\end{tabular}
\endgroup
\caption{Evaluation with different architectures of \textbf{action recognition utility target model} $f'_{T}$. Results shows Top-1 Accuracy (\%) on UCF101. Goal of this evaluation is to maintain the action recognition performance close to the raw data baseline regardless of choice of model $f'_{T}$. Our self-supervised method achieves \textbf{model-agnostic action recognition performance} which is also comparable to the supervised method~\cite{wu_tpami}.}
\label{table:f't}

\end{table}

\subsection{Evaluating on different privacy target model $f'_{B}$}
A learned anonymization function $f^{*}_{A}$ is expected to provide protection against privacy leakage from any privacy target model $f'_{B}$. In training of anonymizatoin function , we use ResNet50 as the auxiliary privacy model $f_{B}$ and evaluate the learned anonymization $f^{*}_{A}$ on target privacy classifiers $f'_{B}$ like ResNet18/50/34/101/152 and MobileNet-V1 with and without ImageNet~\cite{deng2009imagenet} pretraining. From Table~\ref{table:f'b1}, we can observe that our method protects privacy leakage regardless of choice of target privacy model. Using ImageNet pretraining as shown in Table ~\ref{table:f'b2}, privacy leakage increases in all methods, however, the relative drop to the raw data baseline is improved.

\begin{table*}[h]
\small
\centering
\begingroup
\setlength{\tabcolsep}{2pt} %
\renewcommand{\arraystretch}{1.0}
\begin{tabular}{lcccccccccccc} 
\hline

\hline

\hline\\[-3mm]
\multirow{2}{*}{Method} & \multicolumn{2}{c}{\textbf{ResNet18}} & \multicolumn{2}{c}{\textbf{ResNet34}} & \multicolumn{2}{c}{\textbf{ResNet50}} & \multicolumn{2}{c}{\textbf{ResNet101}} & \multicolumn{2}{c}{\textbf{\textbf{ResNet152}}} & \multicolumn{2}{c}{\textbf{MobileNet-V1}}  \\
                        & cMAP (\%)($\downarrow$) & F1($\downarrow$)               & cMAP (\%) & F1               & cMAP (\%) & F1               & cMAP (\%) & F1                & cMAP (\%) & F1                & cMAP (\%) & F1                    \\ 
\hline
Raw data                & 64.38     & 0.5385           & 65.30     & 0.5554           & 64.40     & 0.5553           & 60.70     & 0.5269            & 58.83     & 0.4852            & 61.21     & 0.5056                \\
Supervised              & 53.84     & 0.4402           & 53.22     & 0.4283           & 53.97     & 0.4459           & 53.55     & 0.4257            & 51.05     & 0.4030            & 52.48     & 0.4013                \\
\textbf{Ours}                    & 54.83     & 0.4574           & 54.09     & 0.4226           & 57.43     & 0.4732           & 52.94     & 0.4096            & 53.27     & 0.4322            & 53.41     & 0.3974                \\
\hline

\hline

\hline\\[-3mm]
\end{tabular}
\endgroup
\caption{Evaluating $f^{*}_{A}$ for privacy leakage against different architectures of \textbf{privacy target model} $f'_{B}$. Results shown on VISPR-1 dataset. Lower privacy classification score is better. Our self-supervised method gets a \textbf{model-agnostic privacy anonymization performance} which is also comparable to the supervised method~\cite{wu_tpami}.}
\label{table:f'b1}

\end{table*}

\begin{table*}
\centering
\small
\begingroup
\setlength{\tabcolsep}{2pt} %
\renewcommand{\arraystretch}{1.0}
\begin{tabular}{lcccccccccc} 
\hline

\hline

\hline\\[-3mm]
\multirow{2}{*}{Method} & \multicolumn{2}{c}{\textbf{ResNet18}} & \multicolumn{2}{c}{\textbf{ResNet34}} & \multicolumn{2}{c}{\textbf{ResNet50}} & \multicolumn{2}{c}{\textbf{ResNet101}} & \multicolumn{2}{c}{\textbf{ResNet152}}  \\
                        & cMAP (\%) & F1                        & cMAP (\%) & F1                        & cMAP (\%) & F1                        & cMAP (\%) & F1                         & cMAP (\%) & F1                          \\ 
\hline
Raw data                & 69.82     & 0.6041                    & 69.55     & 0.6447                    & 70.66     & 0.6591                    & 71.09     & 0.6330                     & 69.50     & 0.6130                      \\
Supervised              & 58.05     & 0.5367                    & 58.02     & 0.5463                    & 62.01     & 0.5281                    & 61.44     & 0.5553                     & 61.88     & 0.5711                      \\
\textbf{Ours}                    & 59.10     & 0.5302                    & 59.71     & 0.5227                    & 60.73     & 0.5689                    & 59.24     & 0.5601                     & 60.51     & 0.5352                      \\
\hline

\hline

\hline\\[-3mm]
\end{tabular}
\endgroup
\caption{ Similar setting as Table~\ref{table:f'b1}, but $f'_{B}$ is initialized with \textbf{ImageNet Pretraining}.}
\label{table:f'b2}

\end{table*}

\subsection{Evaluation protocol: Pretrained Action classifier and fixed privacy classifier}
\begin{figure*}
        \centering
        \includegraphics[width=0.7\textwidth]{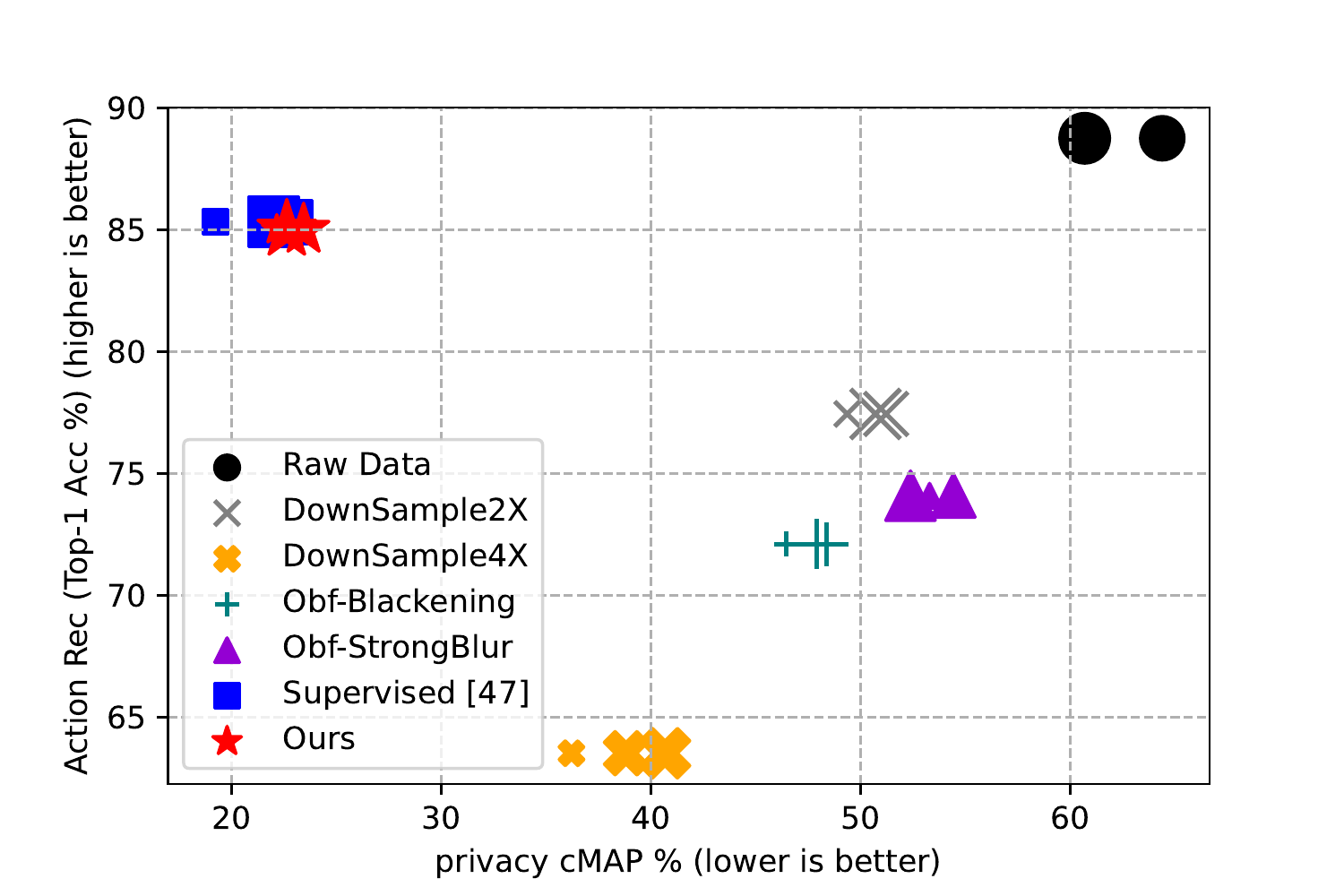}
        \caption{Trade-off between action classification using \textbf{pretrained} action classifier and \textbf{raw-data frozen} privacy classifier. UCF101 is used as action classification dataset and VISPR is used as privacy dataset. Increasing size of the marker shows increasing size of privacy classifiers: ResNet18, ResNet50, ResNet101.}
        \label{fig:bigtradeoff}
\end{figure*}

In a practical scenario, we can initialize an action recognition target model $f'_{T}$ from the Kinetics400 raw data pretrained checkpoint. Also, an intruder has no direct access to the learned anonymization function in a practical setting, hence we can consider the raw-data pretrained privacy classifier as a target privacy model $f'_{B}$. Results are shown in Table~\ref{table:practical_protocol}. We use Kinetics400 pretrained R2Plus1D-18 model as the action recognition target model $f'_{T}$, and ResNet models with varying capacity as the target privacy model $f'_{B}$. Plotting the trade-off of Table~\ref{table:practical_protocol} in Fig.~\ref{fig:bigtradeoff}, we can observe that at the cost of a small drop in action recognition performance our method obtains about \textbf{66\% reduction in privacy leakage} from the raw data baseline. This highlights the potential of our self-supervised privacy preserving framework in a practical scenario without adding cost of privacy annotation in training. 

\begin{table}
\centering
\small
\begingroup
\setlength{\tabcolsep}{2pt} %
\renewcommand{\arraystretch}{1.0}
\begin{tabular}{lcccc} 
\hline

\hline

\hline\\[-3mm]
\multirow{2}{*}{Method} & \multirow{2}{*}{\begin{tabular}[c]{@{}c@{}}Top-1 Acc \\(\%) ($\uparrow$)\end{tabular}} & \multicolumn{3}{c}{cMAP (\%) ($\downarrow$)}    \\
                        &                                                                           & \textbf{ResNet18} & \textbf{ResNet50} & \textbf{ResNet101}  \\ 
\hline
Raw data                & 88.76                                                                     & 64.38    & 64.40    & 60.70      \\
Downsample-2x           & 77.45                                                                     & 49.37    & 51.23    & 50.72      \\
Downsample-4x           & 63.53                                                                     & 36.22    & 38.82    & 40.68      \\
Obf-Blackening          & 72.11                                                                     & 46.48    & 48.38    & 47.92      \\
Obf-StrongBlur          & 74.10                                                                     & 53.30    & 54.44    & 52.39      \\
Supervised              & 85.33                                                                     & 19.23\increase{70\%}    & 22.81\increase{64\%}    & 22.01\increase{64\%}      \\
\textbf{Ours}                    & 85.01                                                                     & 22.16\increase{66\%}    & 23.44\increase{64\%}    & 22.64\increase{63\%}      \\
\hline

\hline

\hline\\[-3mm]
\end{tabular}
\endgroup
\caption{Trade-off between action classification and privacy classifier in a \textbf{practical scenario} where target utility model is taken from Kinetics400 checkpoint and target privacy model is raw-data pretrained. UCF101 is used as action classification dataset and VISPR is used as privacy dataset. \increase{\%} denotes relative drop from raw data. With a small drop in action recognition performance our method greatly reduce privacy leakage.}
\label{table:practical_protocol}
\end{table}

\subsection{Plots for known and novel action and privacy attributes protocol}
\begin{figure*}
\centering
    \begin{subfigure}{0.49\textwidth}
    \centering

          \includegraphics[width=\columnwidth]{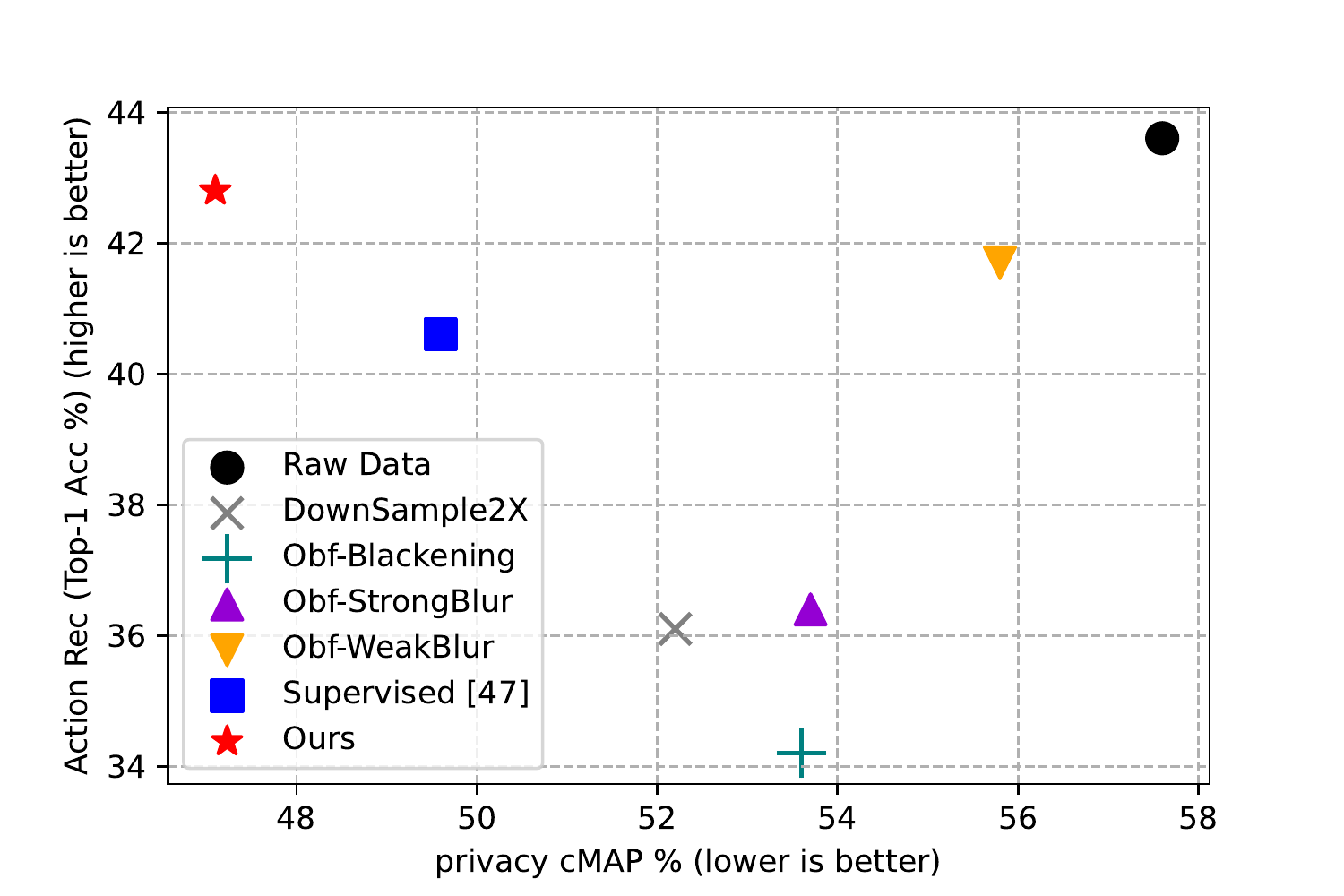}

        \caption{Trade-off between action classification and privacy removal while generalizing from \textbf{UCF101$\rightarrow$PA-HMDB} for action and \textbf{VISPR1$\rightarrow$VISPR2} for privacy attributes.}
    \end{subfigure}
    \hfill
    \begin{subfigure}{0.49\textwidth}
        \centering
  \includegraphics[width=\columnwidth]{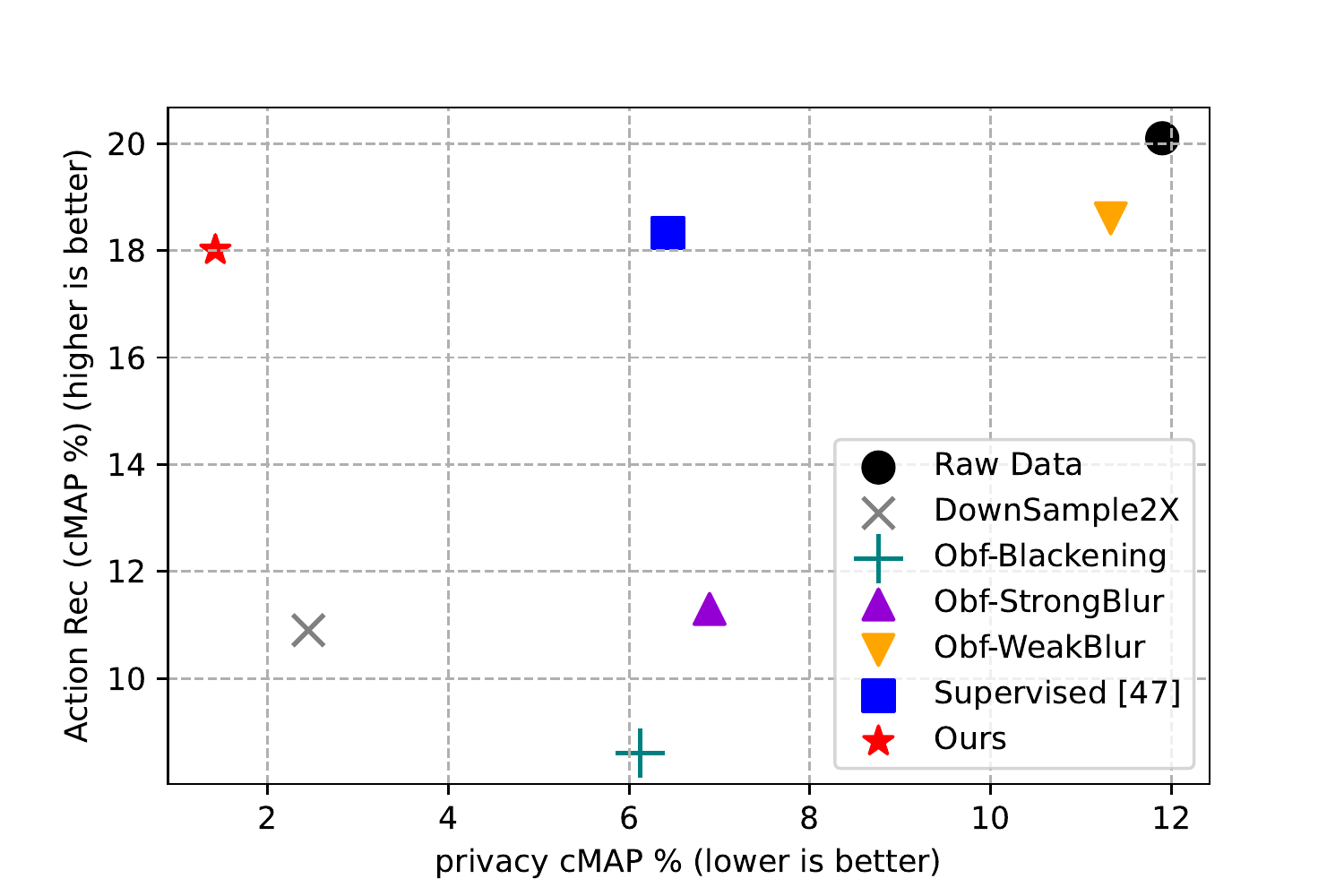}
\caption{Trade-off between action classification and privacy removal while generalizing from \textbf{Scenes$\rightarrow$Objects} for privacy attributes on \textbf{P-HVU} dataset.}
    \end{subfigure}
    
    \caption{Evaluating learned anonymization for \textbf{novel action-privacy attributes}. Our framework outperforms the supervised method~\cite{wu_tpami} and achieves \textbf{robust generalization} across novel action-privacy attributes. For more details refer \supp{Sec. 5.4} of \supp{main paper}.}    
    \label{fig:novel}
\end{figure*}

\begin{figure*}
\centering
    \begin{subfigure}{0.49\textwidth}
    \centering

        \includegraphics[width=\textwidth]{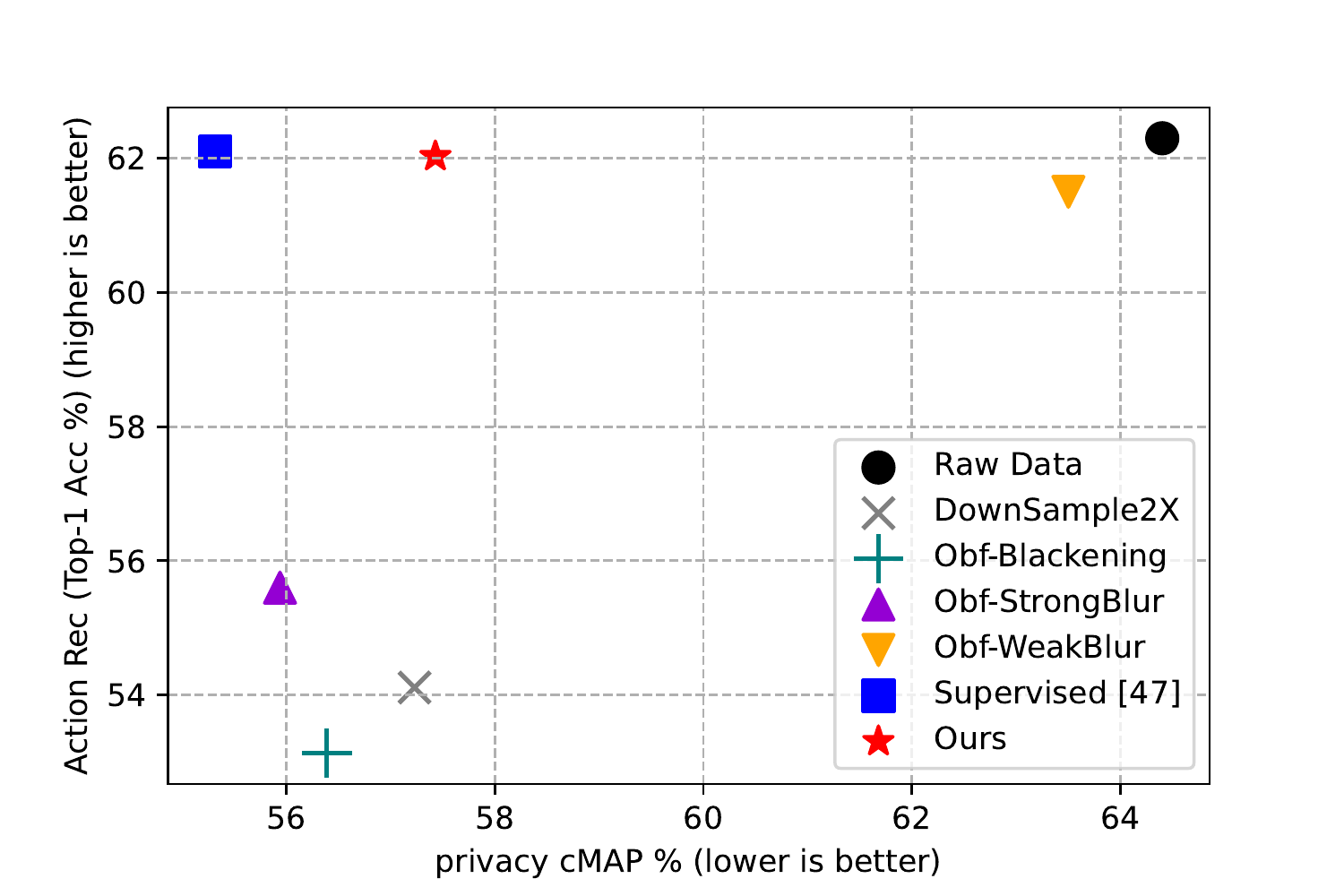}
        \caption{Trade-off between action classification on \textbf{UCF101} vs privacy classification on \textbf{VISPR-1}.}
    \end{subfigure}
    \hfill
    \begin{subfigure}{0.49\textwidth}
        \centering
        \includegraphics[width=\textwidth]{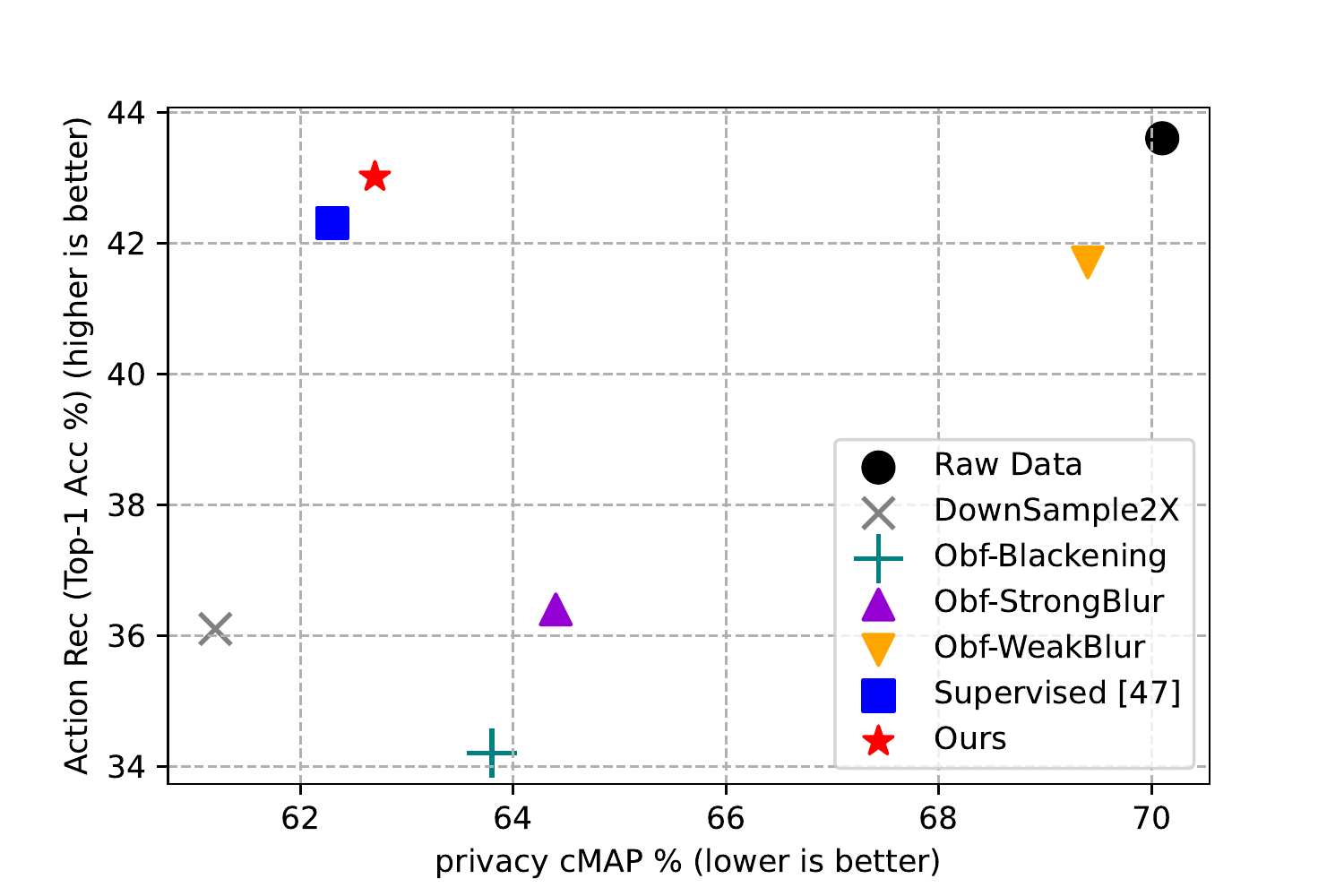}
        \caption{Trade-off between action classification vs privacy classification on \textbf{PA-HMDB}.}
    \end{subfigure}
    \hfill
    \begin{subfigure}{0.49\textwidth}
        \centering
        \includegraphics[width=\columnwidth]{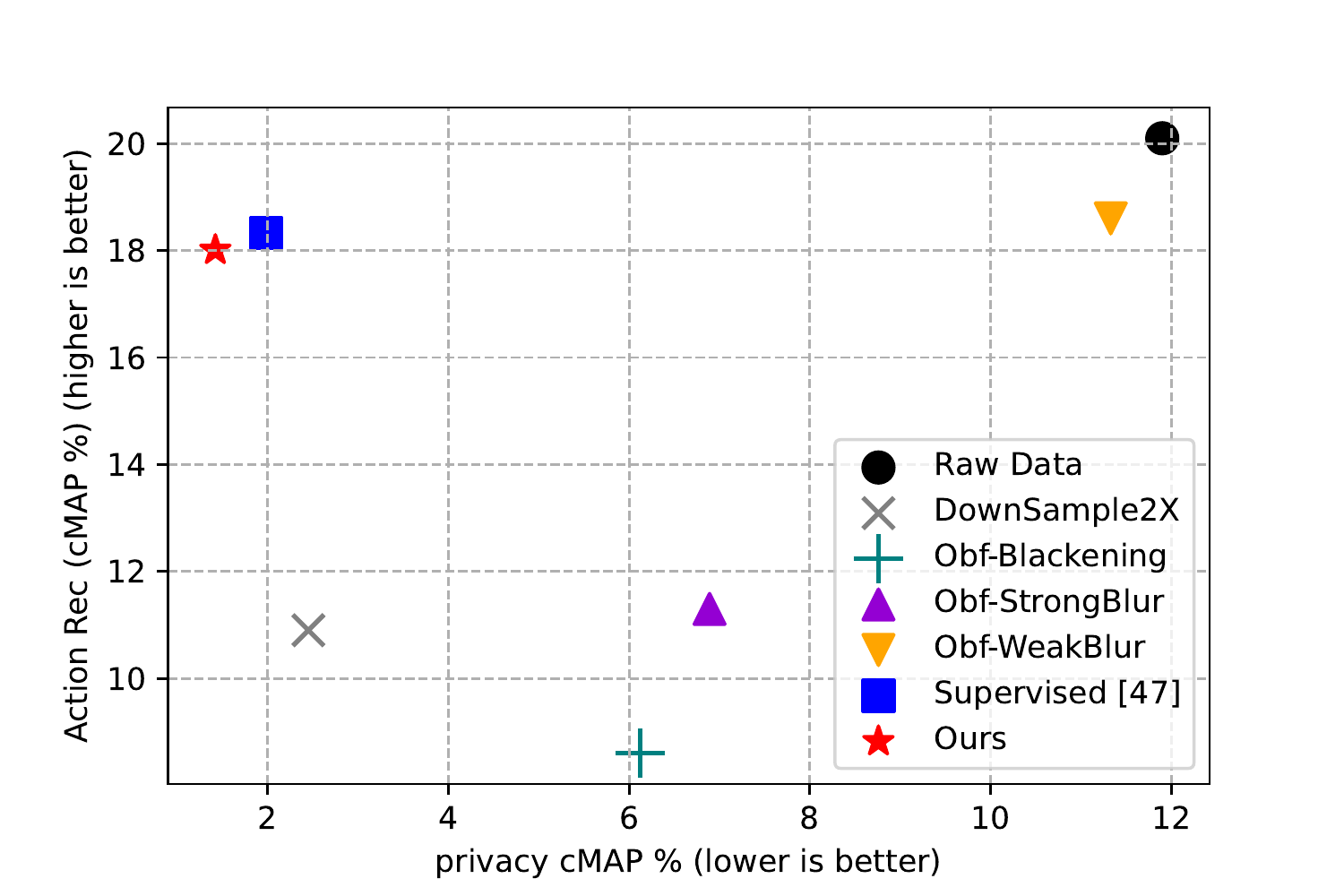}
        \caption{Trade-off between action classification vs \textbf{privacy-object} classification on \textbf{P-HVU}.}
    \end{subfigure}
    \hfill
    \begin{subfigure}{0.49\textwidth}
        \centering
        \includegraphics[width=\columnwidth]{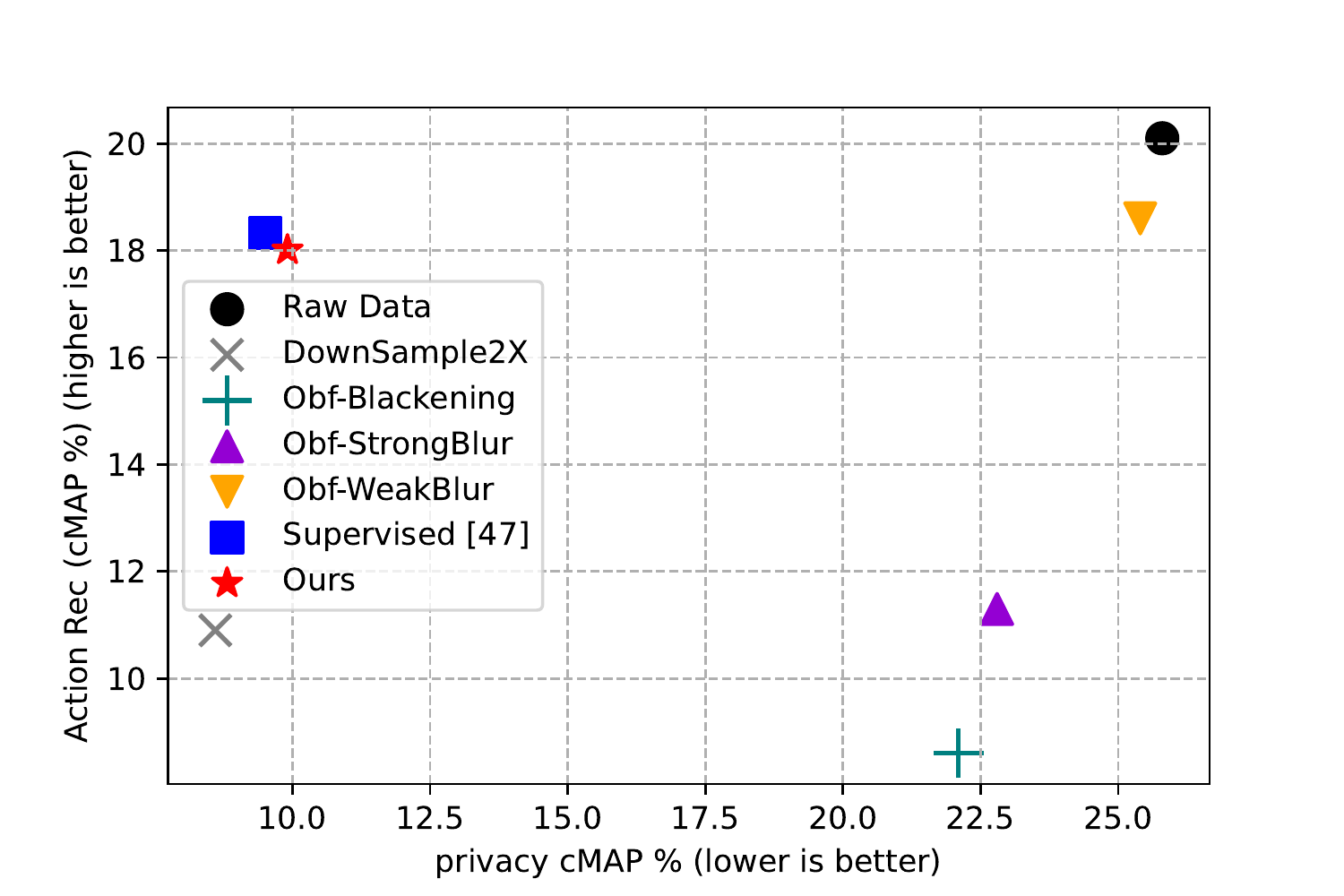}
        \caption{Trade-off between action classification vs \textbf{privacy-scene} classification on \textbf{P-HVU}.}
    \end{subfigure}

    \caption{Evaluating learned anonymization for \textbf{known action-privacy attributes}. Our framework achieves comparable performance to the supervised method~\cite{wu_tpami}. For more details refer \supp{Sec. 5.3} of \supp{main paper}.}    
    \label{fig:known}
\end{figure*}

A trade-off plot for evaluating learned $f^{*}_{A}$ for novel action-privacy attributes is shown in Fig.~\ref{fig:novel} and known action-privacy attributes is shown in Fig~\ref{fig:known}, for more details see \supp{Sec. 5} of \supp{main paper}.

\section{Additional ablations}
\label{sec:addtional_ablation}
\subsection{Effect of different $f_{T}$ architectures} 
To understand the effect of auxiliary model $f_{T}$ in the training process of $f_{A}$, we experiment with different utility auxiliary model $f_{T}$, and report the performance of their learned $f^{*}_{A}$ in the same evaluation setting as shown in Table~\ref{table:ft}. We can observe that there is no significant effect of $f_{T}$ in learning the $f_{A}$. 

\begin{table}
\centering
\small
\begin{tabular}{lccc}
\hline

\hline

\hline\\[-3mm]
\multirow{2}{*}{$f_{T}$ architecture} & UCF101 & \multicolumn{2}{c}{VISPR1}  \\
                                 & Top-1(\%) ($\uparrow$)     & cMAP(\%) ($\downarrow$)   & F1 ($\downarrow$)                \\ 
\hline
R3D-18                           & 62.03  & 57.43 & 0.4732               \\
R2+1D-18                         & 62.37  & 57.37 & 0.4695               \\
R3D-50                           & 62.58  & 57.51 & 0.4707               \\
\hline

\hline

\hline\\[-3mm]
\end{tabular}
\vspace{-3mm}
\caption{\textbf{Auxiliary utility model} $f_{T}$ architecture has no significant effect on final action-privacy measures. Auxiliary models are just used to train the anonymization function and discarded after that. All results are reported on ResNet50 privacy target model $f'_{B}$ and R3D-18 action recognition  target model $f'_{T}$.}
\label{table:ft}
\end{table}

\section{Qualitative Results}
\label{sec:qualitative}
\subsection{Visualization of learned anonymization $f^{*}_{A}$ at different stages of training}
\begin{figure*}[h]
\centering

        \includegraphics[width=\textwidth, trim = 4.1cm 0 4.1cm 0, clip]{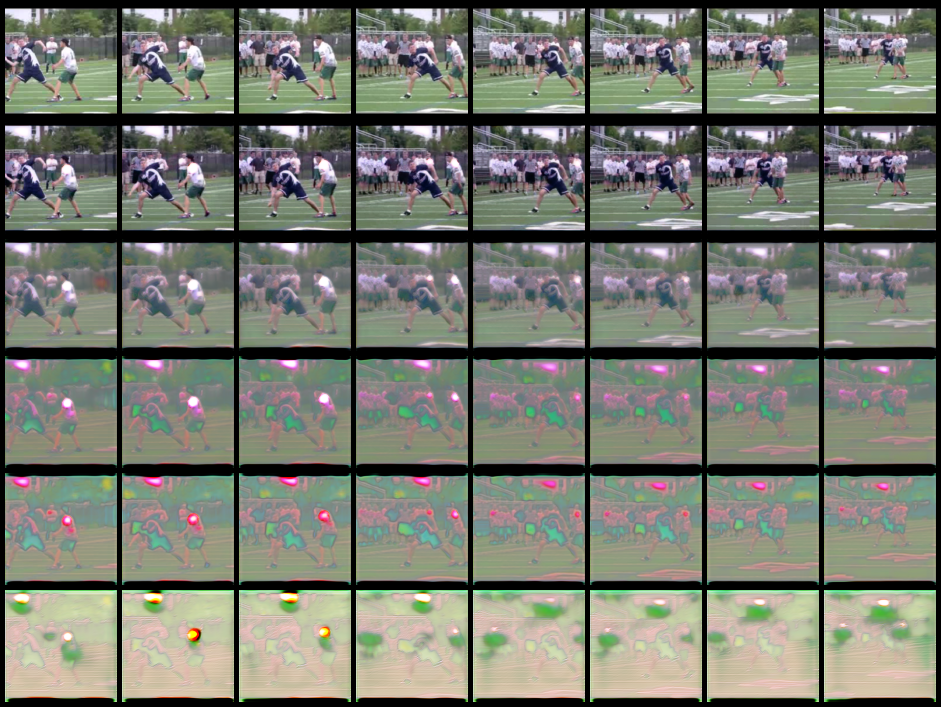}

    \caption{Learned anonymization using our self-supervised privacy preservation framework on test set of UCF101. Groundtruth action label: \texttt{FrisbeeCatch}. First row: original video, from second to last row: anonymized version of video at epoch 1, 3, 6, 9, 30.}    
    \label{fig:epochwise1}
\end{figure*}

\begin{figure*}[h]
\centering

        \includegraphics[width=\textwidth, trim = 4.1cm 0 4.1cm 0, clip]{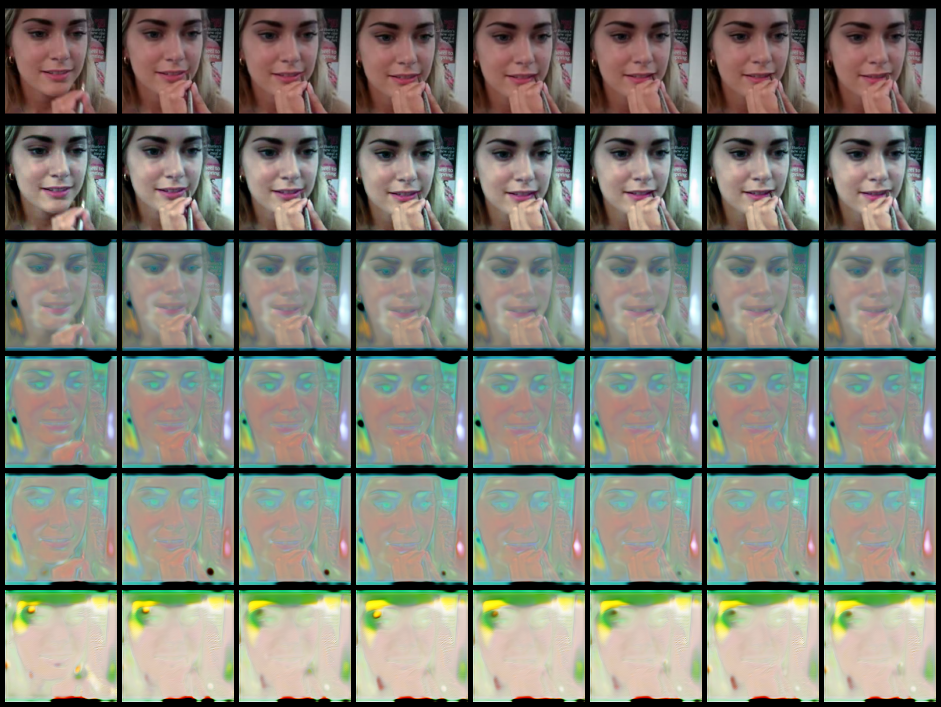}

    \caption{Learned anonymization using our self-supervised privacy preservation framework on test set of UCF101. Groundtruth action label: \texttt{ApplyLipstick}. First row: original video, from second to last row: anonymized version of video at epoch 1, 3, 6, 9, 30.}    
    \label{fig:epochwise2}
\end{figure*}

\begin{figure*}[h]
\centering

        \includegraphics[width=\textwidth, trim = 4.1cm 0 4.1cm 0, clip]{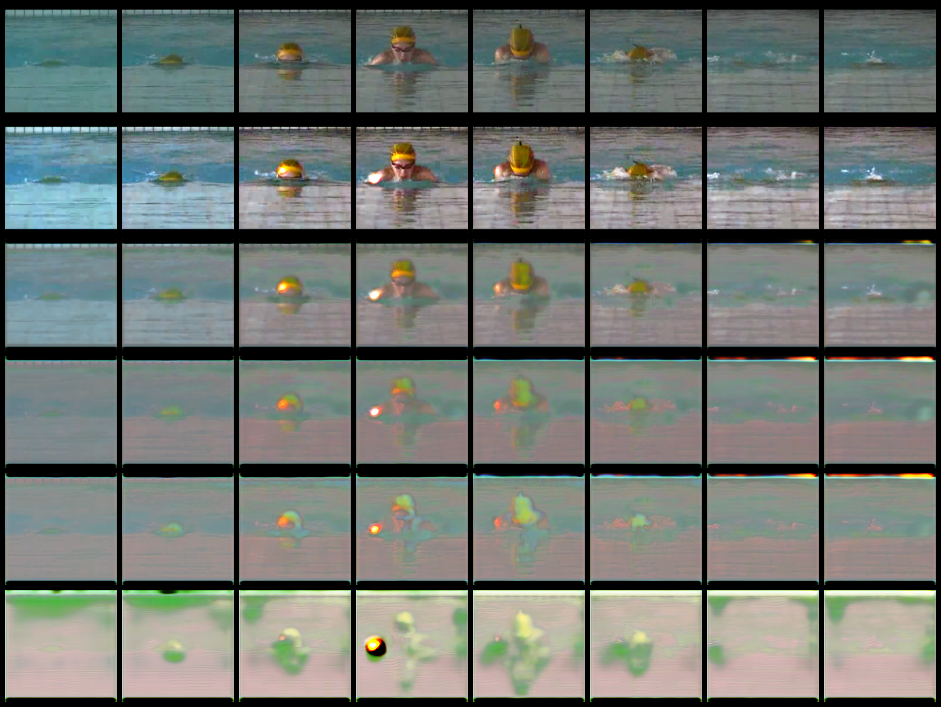}

    \caption{Learned anonymization using our self-supervised privacy preservation framework on test set of UCF101. Groundtruth action label: \texttt{BreastStroke}. First row: original video, from second to last row: anonymized version of video at epoch 1, 3, 6, 9, 30.}    
    \label{fig:epochwise3}
\end{figure*}

In order to visualize the transformation due to learned anonymization function $f^{*}_{A}$, we experiment with various test set videos of UCF101. The sigmoid function after the $f^{*}_{A}$ ensure (0,1) range of the output image. We visualize output at different stages of anonymization training as shown in Fig.~\ref{fig:epochwise1}, ~\ref{fig:epochwise2}, ~\ref{fig:epochwise3}. We can see our self-supervised framework is successfully able to achieve anonymization as the training progresses.

\subsection{Visualization of learned anonymization $f^{*}_{A}$ for different methods}
Apart from Fig.~\ref{fig:epochwise1}, ~\ref{fig:epochwise2}, ~\ref{fig:epochwise3} visualization of our method, we show visualization for all methods, attached in the form of videos in the supplementary zip file.

\subsection{Attention map for supervised vs self-supervised privacy removal branch}
\begin{figure*}[h]
\centering
    \begin{subfigure}{\textwidth}
    \centering

        \includegraphics[width=0.7\textwidth, trim = 9.6cm 0 9.6cm 0, clip]{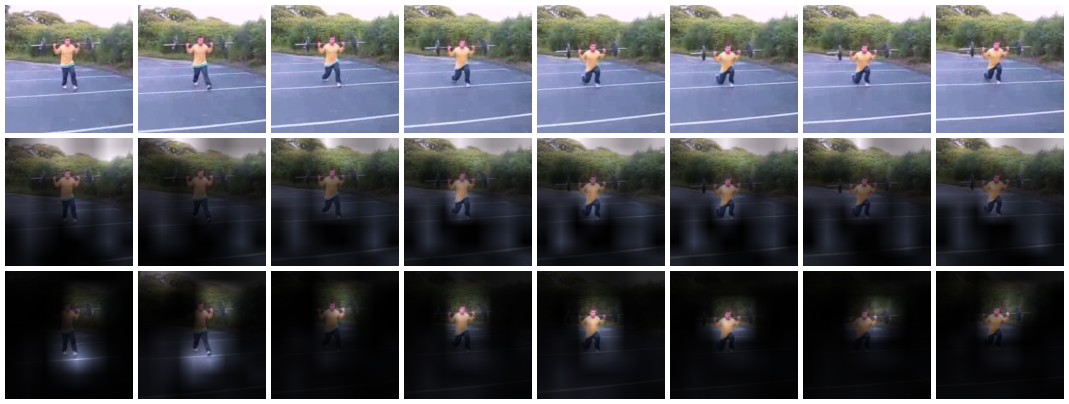}
        \caption{\texttt{Lunges}}
    \end{subfigure}
    \begin{subfigure}{\textwidth}
    \centering

        \includegraphics[width=0.7\textwidth, trim = 9.6cm 0 9.6cm 0, clip]{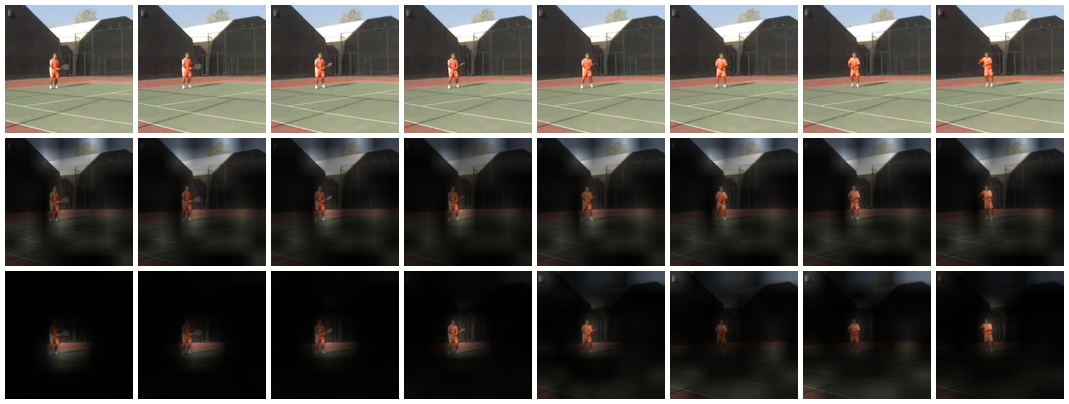}
        \caption{\texttt{TennisSwing}}
        \end{subfigure}

    \vspace{-3mm}
    \caption{\textbf{Attention map visualization}: Top row: original video, middle-row: attention of a self-supervised model, bottom-row: attention of supervised privacy classifier. It can be observed that supervised privacy classifier mainly focuses on the semantics of human, whereas self-supervised model learns holistic spatial semantic features related to the \textbf{scene} (eg. \textbf{track-field} in (a) and \textbf{tennis court} in (b)) as well.}     
    \label{fig:att_scene}
\end{figure*}

\begin{figure*}[h]
\centering
\begin{subfigure}{\textwidth}
    \centering

        \includegraphics[width=0.7\textwidth, trim = 9.6cm 0 9.6cm 0, clip]{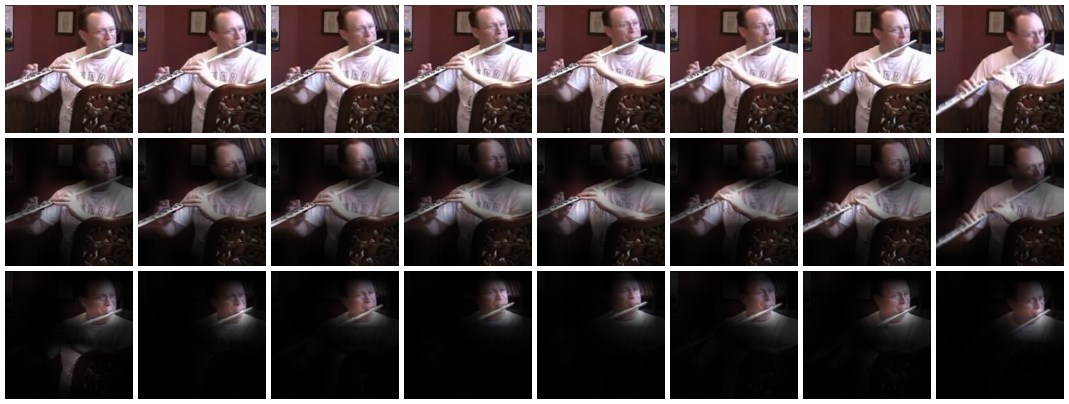}
        \caption{\texttt{PlayingFlute}}
    \end{subfigure}
    \begin{subfigure}{\textwidth}
    \centering

        \includegraphics[width=0.7\textwidth, trim = 9.6cm 0 9.6cm 0, clip]{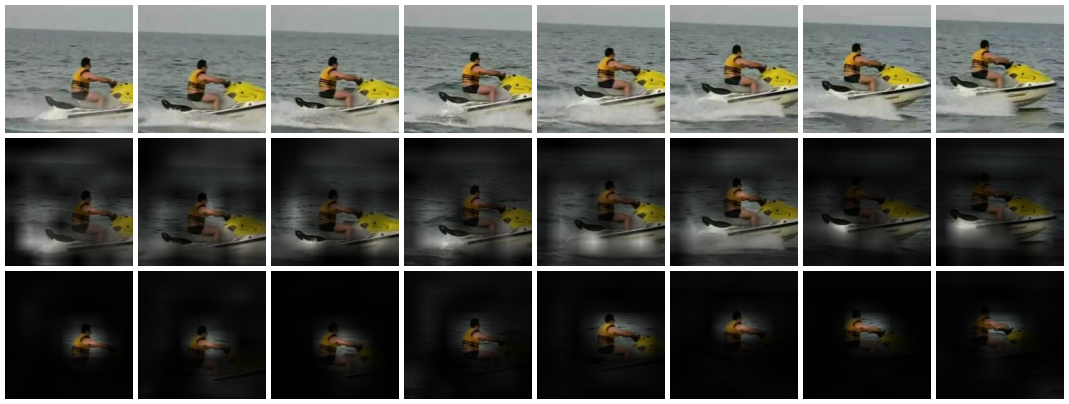}
        \caption{\texttt{Skijet}}
    \end{subfigure}

    \vspace{-3mm}
    \caption{\textbf{Attention map visualization}: Top row: original video, middle-row: attention of a self-supervised model, bottom-row: attention of supervised privacy classifier. It can be observed that supervised privacy classifier mainly learns semantics of human, whereas self-supervised model learns holistic semantic spatial features related to the \textbf{objects} (eg. \textbf{Flute} in (a) and \textbf{SkiJet} in (b)) as well.}    
    \label{fig:att_obj}
\end{figure*}

A self-supervised model focuses on \textbf{holistic spatial semantics}, whereas a supervised privacy classifier focuses on specific semantics of the privacy attributes. To bolster this observation, we visualize the attention map of ResNet50 model which is trained in (1) Supervised manner using binary cross entropy loss using VISPR-1. (2) Self-supervised manner using NT-Xent loss. We use the method of Zagoruyko and  Komodakis~\cite{attentiontransfer} to generate model attention from the third convolutional block of the ResNet model. As can be observed from the attention map visualization of Fig.~\ref{fig:att_scene} that a self-supervised model focuses on semantics related to human and its surrounding \textbf{scene}, whereas, the supervised privacy classifier mainly focuses on the human semantics. In Fig.~\ref{fig:att_obj}, we can see that the self-supervised model attends to the semantics of \textbf{object} along with human, and supervised privacy classifier mainly learns semantics of human only. 

\section{Visual Aid for training and evaluation protocols}
\label{sec:protocol}
\begin{figure*}[h]
\centering
    \begin{subfigure}{\textwidth}
    \centering

        \includegraphics[width=0.6\textwidth]{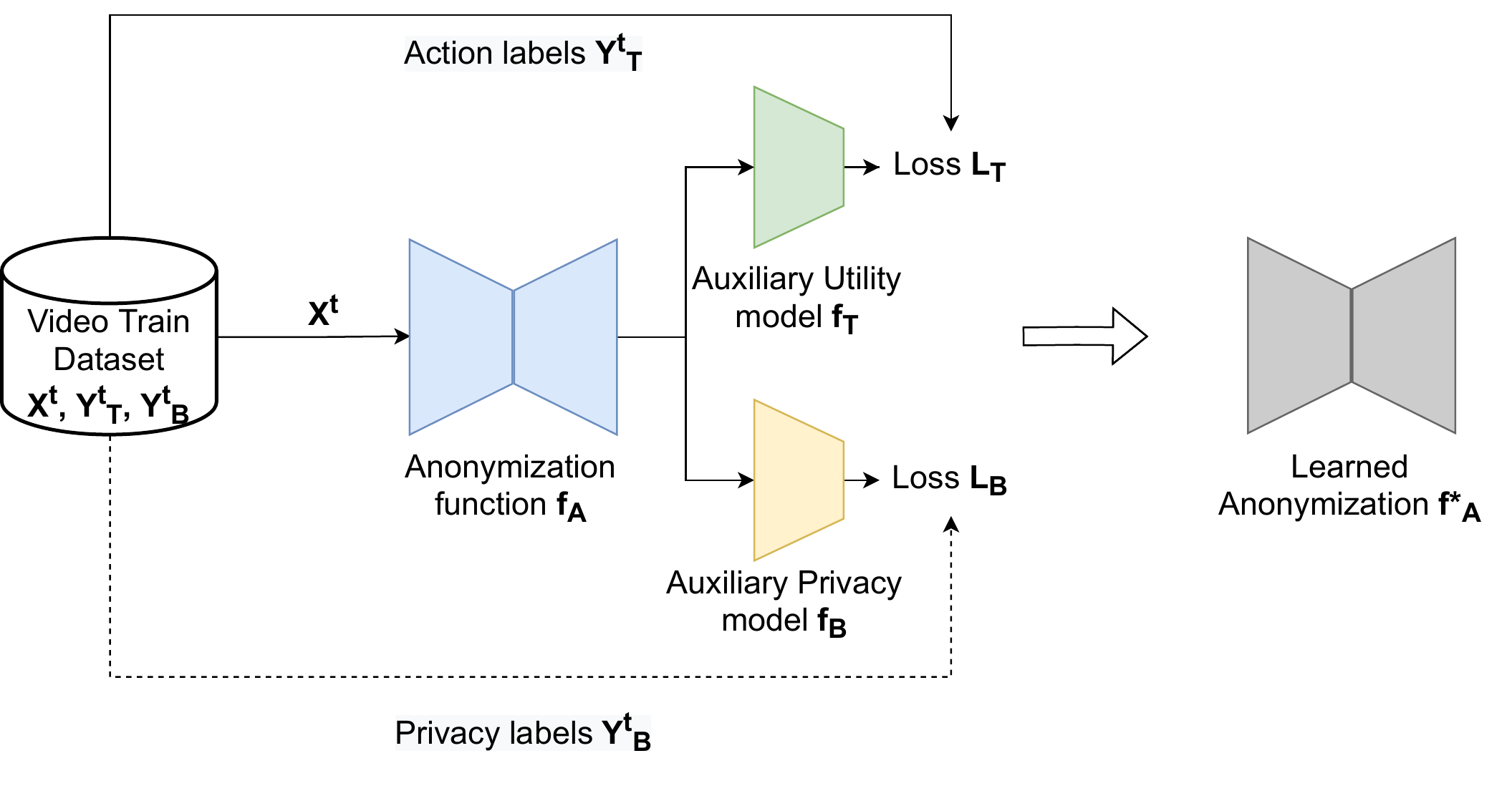}
        \caption{\textbf{\ul{First phase}: Training of anonymization function $f_{A}$}. For our self-supervised method we do not require privacy labels $Y^{t}_{B}$. At the end of training, $f_{A}$ is frozen call it $f^{*}_{A}$, and auxiliary models $f_{B}$ and $f_{T}$ are discarded.}
    \end{subfigure}
    
    \begin{subfigure}{\textwidth}
        \centering
        \includegraphics[width=0.6\textwidth]{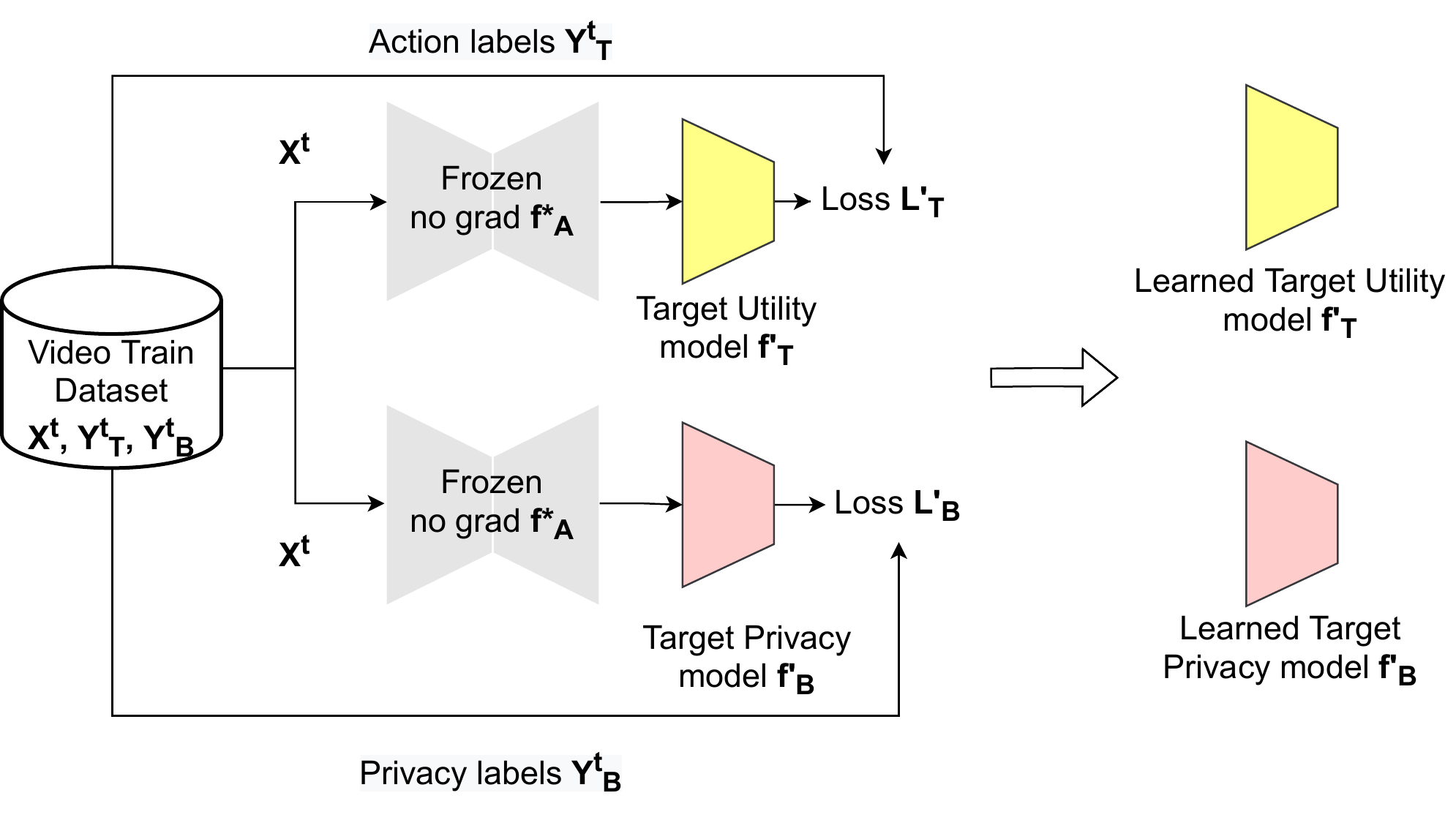}
        \caption{\textbf{\ul{Second phase} :Target models training} Target models are used to evaluate the performance of learned anonymization function $f^{*}_{A}$ and are different from auxiliary models. Target utility model i.e. action classifier $f'_{T}$ and Target privacy classifier $f'_{B}$ are learned in supervised manner on the anonymized version of training data $X^{t}$.}
    \end{subfigure}
    
    \begin{subfigure}{\textwidth}
        \centering
        \includegraphics[width=0.5\textwidth]{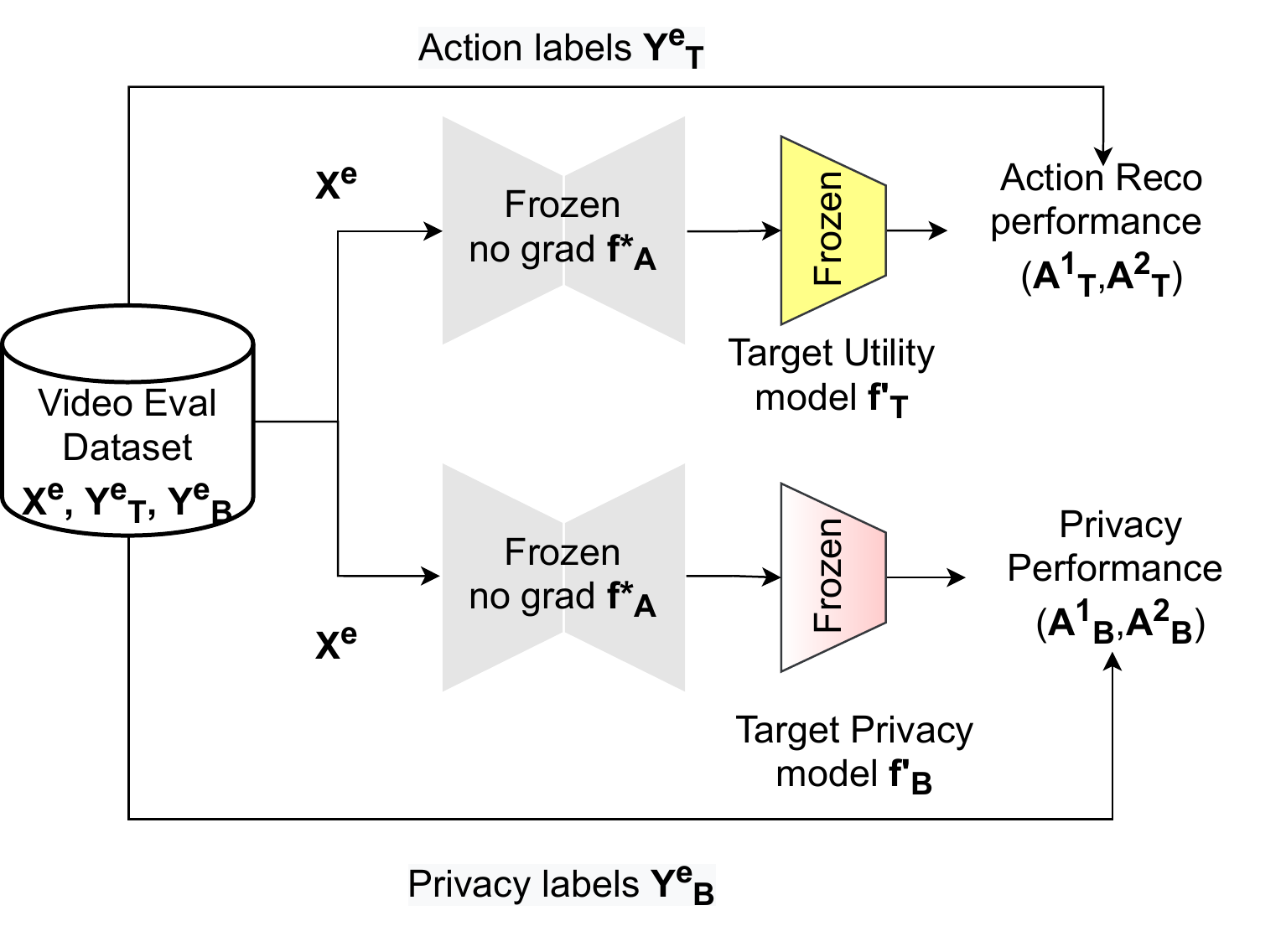}
        \caption{\textbf{\ul{Third phase}: Target models testing}: Once target models are trained from anonymized version of $X^{t}$, they are are frozen and evaluated on anonymized version of test/evaluation set $X^{e}$.}
    \end{subfigure}
    
    \caption{Visual Aid for \textbf{Same-dataset} training and evaluation protocol \supp{Sec. 4.1} of \supp{main paper}.}    
    \label{fig:same_dataset_protocol}
\end{figure*}

\begin{figure*}[h]
\centering
    \begin{subfigure}{\textwidth}
    \centering

        \includegraphics[width=0.6\textwidth]{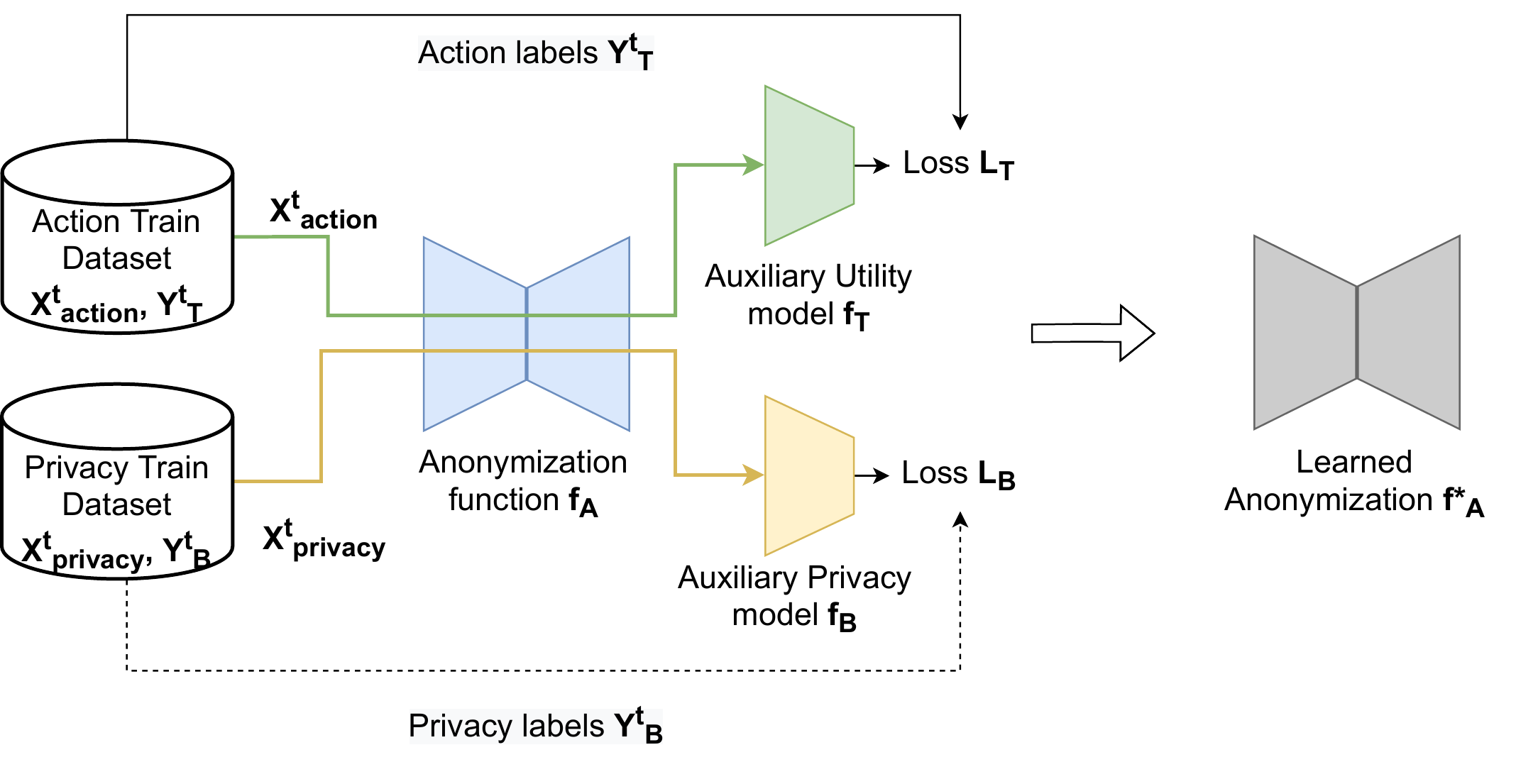}
        \caption{\textbf{\ul{First phase}: Training of anonymization function $f_{A}$}. For our self-supervised method we do not require privacy labels $Y^{t}_{B}$. At the end of training, $f_{A}$ is frozen call it $f^{*}_{A}$, and auxiliary models $f_{B}$ and $f_{T}$ are discarded.}
    \end{subfigure}
    
    \begin{subfigure}{\textwidth}
        \centering
        \includegraphics[width=0.6\textwidth]{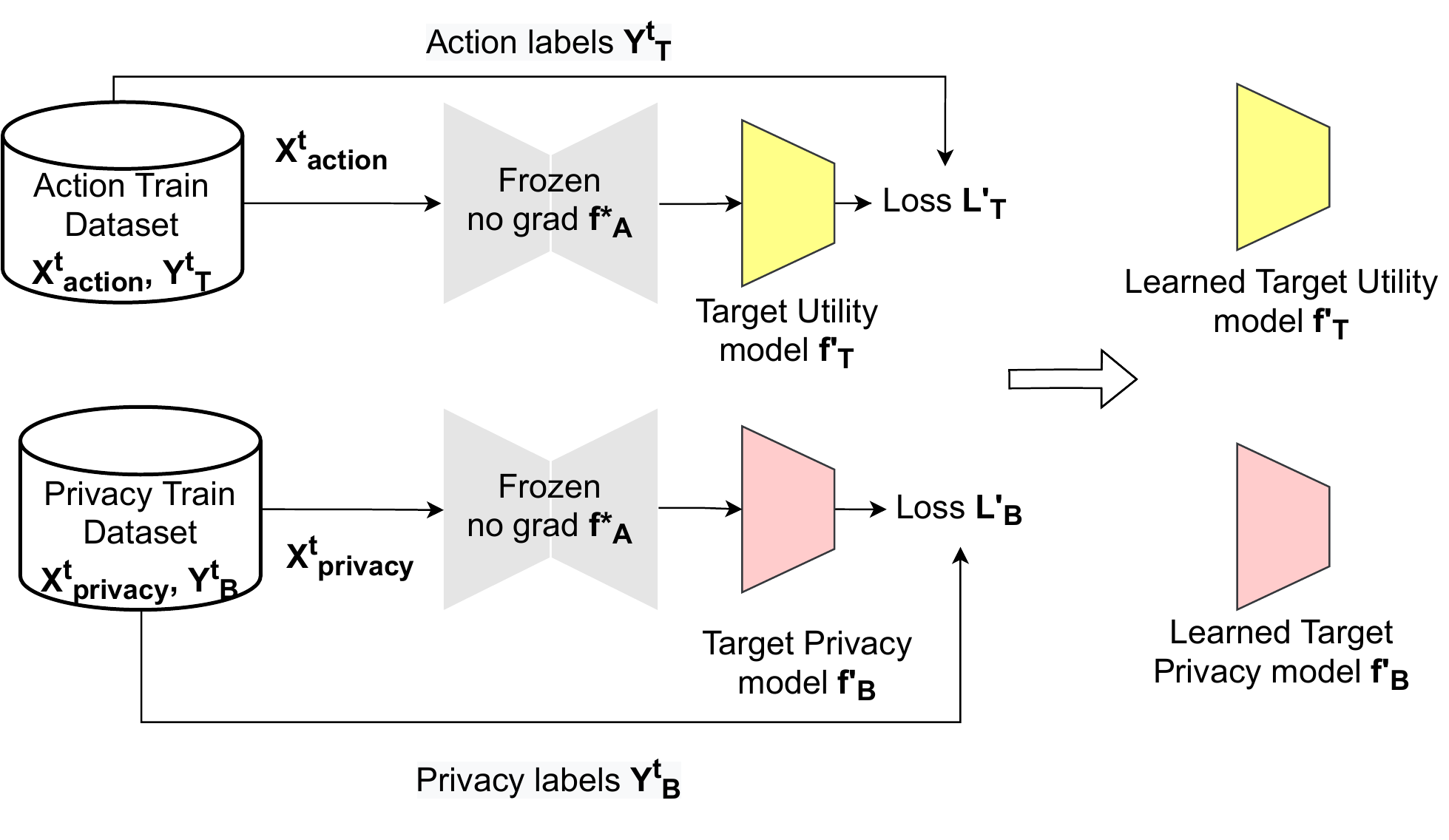}
        \caption{\textbf{\ul{Second phase} :Target models training} Target models are used to evaluate the performance of learned anonymization function $f^{*}_{A}$ and are different from auxiliary models. Target utility model (action classifier) $f'_{T}$ and Target privacy classifier $f'_{B}$ are learned in supervised manner on the anonymized version of training data $X^{t}_{action}$ and $X^{t}_{privacy}$.}
    \end{subfigure}
    
    \begin{subfigure}{\textwidth}
        \centering
        \includegraphics[width=0.5\textwidth]{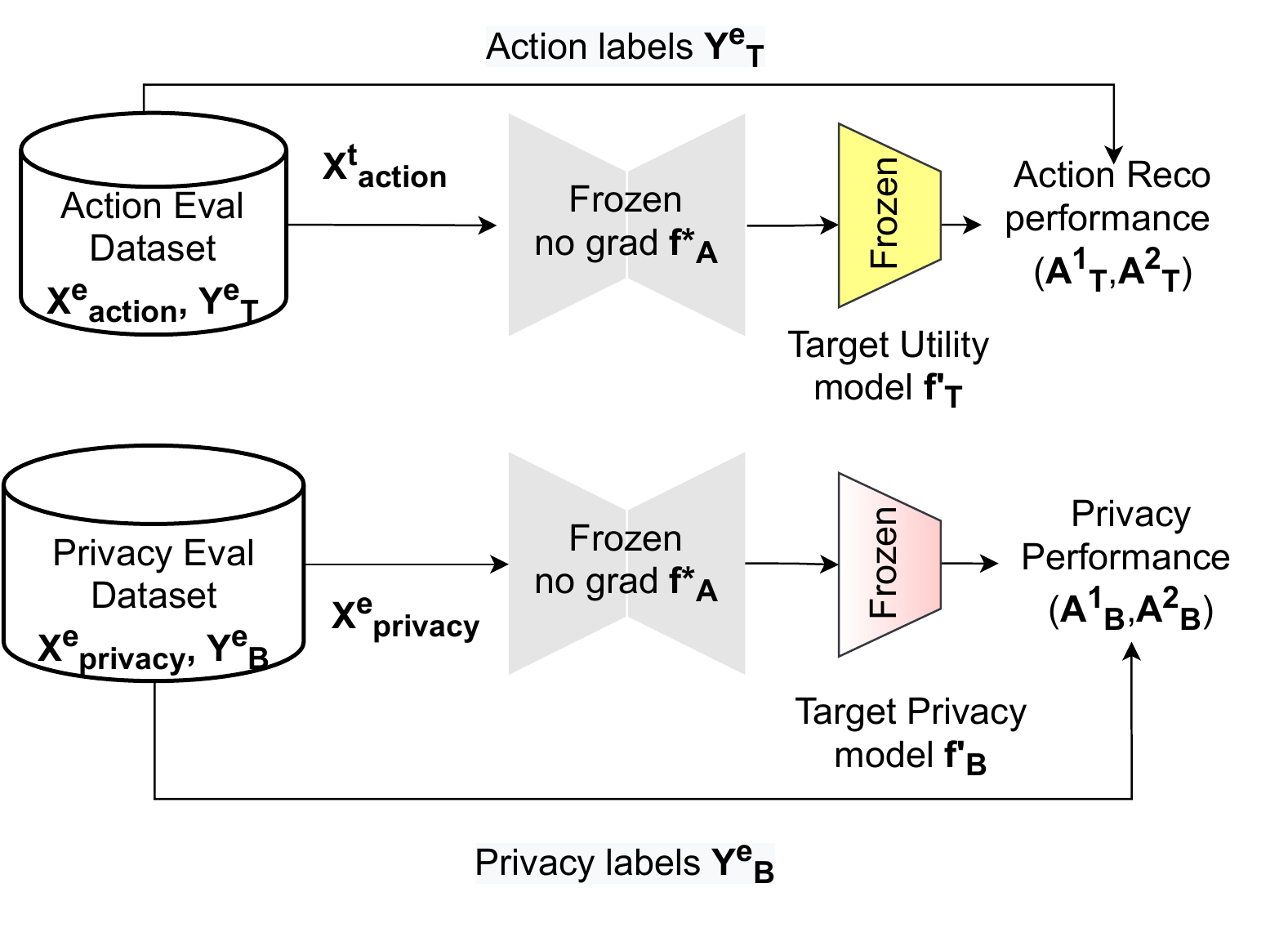}
        \caption{\textbf{\ul{Third phase}: Target models testing}: Once target models are trained from anonymized version of $X^{t}_{action}$, $X^{t}_{privacy}$, they are are frozen and evaluated on anonymized version of test/eval set $X^{e}_{action}$ , $X^{e}_{privacy}$.}
    \end{subfigure}
    
    \caption{Visual Aid for \textbf{Cross-dataset} training and evaluation protocol \supp{Sec. 4.2} of \supp{main paper}.}    
    \label{fig:cross_dataset_protocol}
\end{figure*}

\begin{figure*}[h]
\centering
    \begin{subfigure}{\textwidth}
    \centering

        \includegraphics[width=0.6\textwidth]{figures/supp_figures/visual_aid_set1_1.drawio.pdf}
        \caption{\textbf{\ul{First phase}: Training of anonymization function $f_{A}$}. For our self-supervised method we do not require privacy labels $Y^{t}_{B}$. At the end of training, $f_{A}$ is frozen call it $f^{*}_{A}$, and auxiliary models $f_{B}$ and $f_{T}$ are discarded.}
    \end{subfigure}
    
    \begin{subfigure}{\textwidth}
        \centering
        \includegraphics[width=0.6\textwidth]{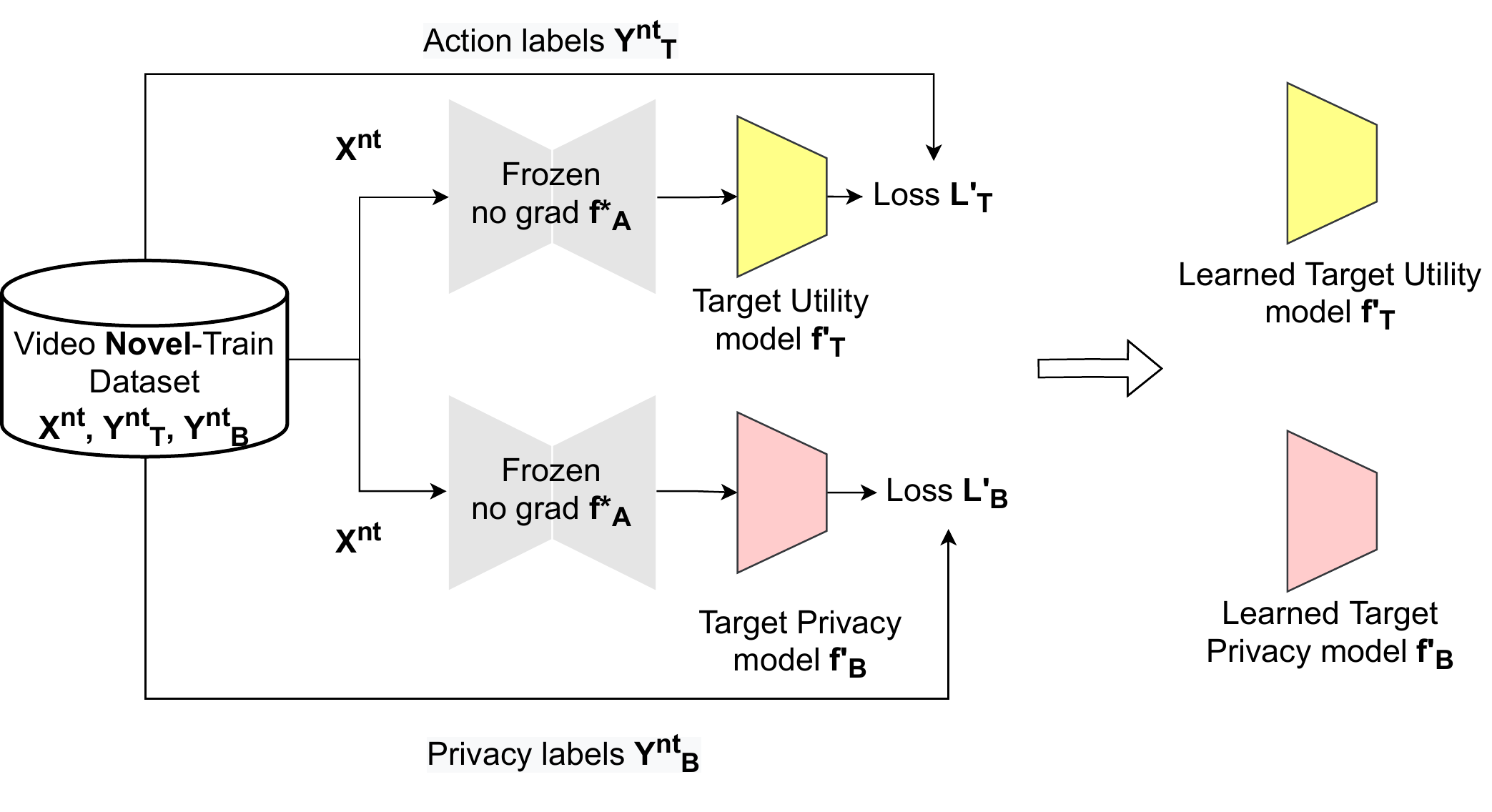}
        \caption{\textbf{\ul{Second phase} :Target models training} Target models are used to evaluate the performance of learned anonymization function $f^{*}_{A}$ and are different from auxiliary models. Target utility model (action classifier) $f'_{T}$ and Target privacy classifier $f'_{B}$ are learned in supervised manner on the anonymized version of \textbf{novel training data} $X^{nt}$ which has action and privacy labels such that $Y^{nt}_{T}\cap Y^{t}_{T}= \phi$ and $Y^{nt}_{B}\cap Y^{t}_{B}= \phi$}
    \end{subfigure}
    
    \begin{subfigure}{\textwidth}
        \centering
        \includegraphics[width=0.5\textwidth]{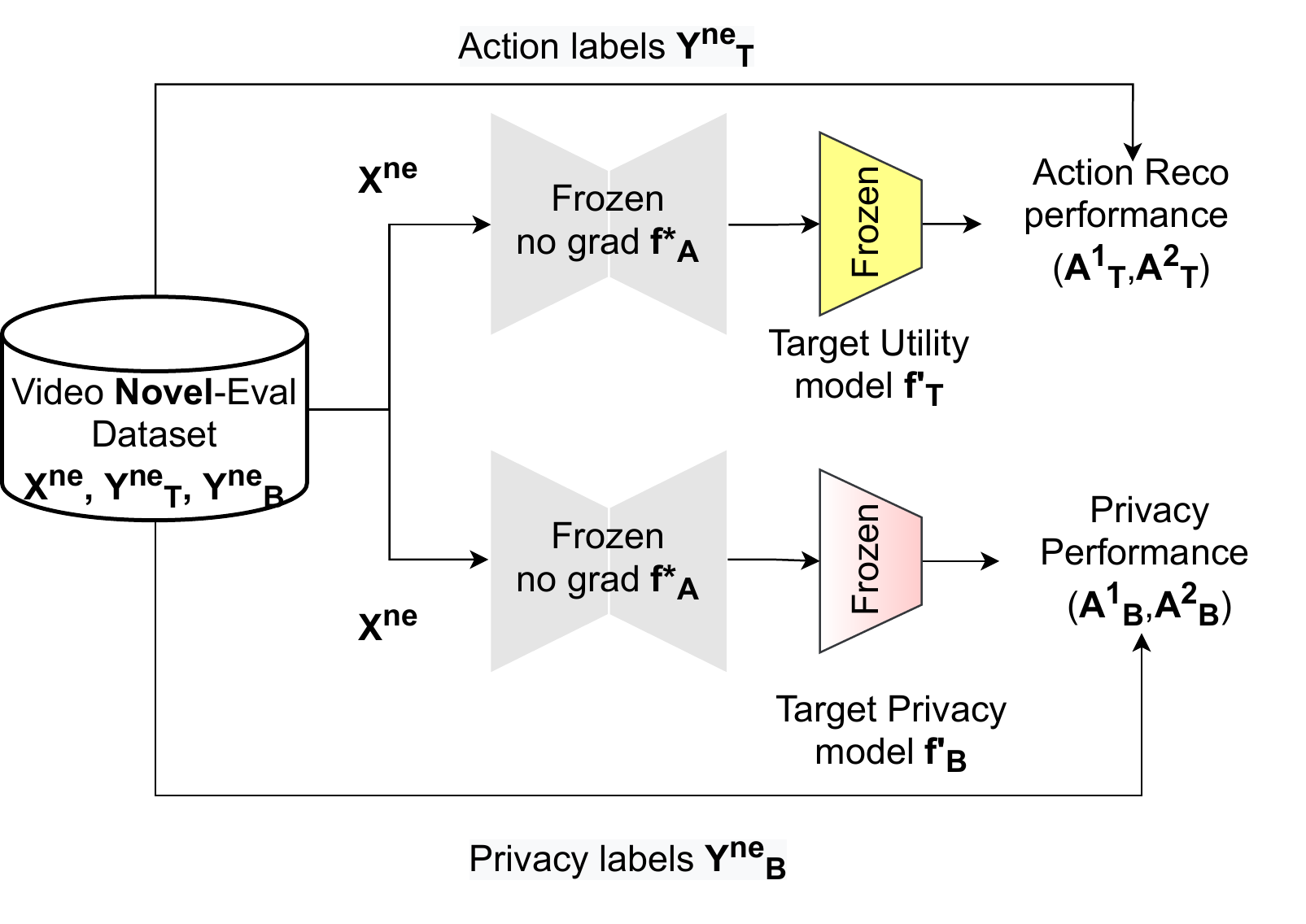}
        \caption{\textbf{\ul{Third phase}: Target models testing}: Once target models are trained from anonymized version of \textbf{novel training data} $X^{nt}$, they are are frozen and evaluated on anonymized version of \textbf{novel test/eval set} $X^{ne}$.}
    \end{subfigure}
    
    \caption{Visual Aid for \textbf{Novel Action and privacy attribution} protocol \supp{Sec. 4.3} of \supp{main paper}.}    
    \label{fig:novel_datseat_protocol}
\end{figure*}

In order to better understand protocols of \supp{Sec. 4} of \supp{main paper}, we provide here some visual aids in Fig~\ref{fig:same_dataset_protocol}, ~\ref{fig:cross_dataset_protocol}, and ~\ref{fig:novel_datseat_protocol}.

\end{document}